\documentclass[sn-mathphys]{sn-jnl}% Math and Physical Sciences Reference Style
\jyear{2021}%

\theoremstyle{thmstyleone}%
\usepackage{multirow}
\usepackage{booktabs,tabularx, array}
\usepackage{wrapfig}
\usepackage{colortbl}

\usepackage{setspace}

\theoremstyle{thmstyletwo}%

\theoremstyle{thmstylethree}%

\begin{document}
% \onehalfspacing % Set line spacing to 1.5

\title[Extreme Image Transformations]{Extreme Image Transformations Affect Humans and Machines Differently}
% possible titles
% Extreme structure distortions for object recognition
% Extreme adversarial attacks on object recognition
% Extreme 
% keywords: extreme structure recognition humans machines

%%=============================================================%%
%% Prefix	-> \pfx{Dr}
%% GivenName	-> \fnm{Joergen W.}
%% Particle	-> \spfx{van der} -> surname prefix
%% FamilyName	-> \sur{Ploeg}
%% Suffix	-> \sfx{IV}
%% NatureName	-> \tanm{Poet Laureate} -> Title after name
%% Degrees	-> \dgr{MSc, PhD}
%% \author*[1,2]{\pfx{Dr} \fnm{Joergen W.} \spfx{van der} \sur{Ploeg} \sfx{IV} \tanm{Poet Laureate} 
%%                 \dgr{MSc, PhD}}\email{iauthor@gmail.com}
%%=============================================================%%

\author[]{\fnm{Girik} \sur{Malik}}%\email{iauthor@gmail.com}

\author[]{\fnm{Dakarai} \sur{Crowder}}%\email{iiauthor@gmail.com}
% \equalcont{These authors contributed equally to this work.}

\author[]{\fnm{Ennio} \sur{Mingolla}}%\email{iiiauthor@gmail.com}
% \equalcont{These authors contributed equally to this work.}

% \affil*[1]{\orgdiv{Department}, \orgname{Organization}, \orgaddress{\street{Street}, \city{City}, \postcode{100190}, \state{State}, \country{Country}}}

\affil[]{\orgname{Northeastern University}, \orgaddress{\city{Boston}, \postcode{02115}, \state{MA}, \country{USA}}}
\affil{\{malik.gi, crowder.d, e.mingolla\}@northeastern.edu}

% \affil[2]{\orgdiv{Department}, \orgname{Organization}, \orgaddress{\street{Street}, \city{City}, \postcode{10587}, \state{State}, \country{Country}}}

% \affil[3]{\orgdiv{Department}, \orgname{Organization}, \orgaddress{\street{Street}, \city{City}, \postcode{610101}, \state{State}, \country{Country}}}

%%==================================%%
%% sample for unstructured abstract %%
%%==================================%%

\abstract{   
Some recent artificial neural networks (ANNs) claim to model aspects of primate neural and human performance data. Their success in object recognition is, however, dependent on exploiting low-level features for solving visual tasks in a way that humans do not. As a result, out-of-distribution or adversarial input is often challenging for ANNs. Humans instead learn abstract patterns and are mostly unaffected by many extreme image distortions. We introduce a set of novel image transforms inspired by neurophysiological findings and evaluate humans and ANNs on an object recognition task. We show that machines perform better than humans for certain transforms and struggle to perform at par with humans on others that are easy for humans. We quantify the differences in accuracy for humans and machines and find a ranking of difficulty for our transforms for human data. We also suggest how certain characteristics of human visual processing can be adapted to improve the performance of ANNs for our difficult-for-machines transforms. 
}


%%\pacs[JEL Classification]{D8, H51}

%%\pacs[MSC Classification]{35A01, 65L10, 65L12, 65L20, 65L70}

\maketitle

\section{Introduction}
\label{sec:intro}
% basic motivation
% talk about the meaning of the object for humans and machines
% link with Ullman's PNAS paper, learning with noise and the use of additional blocks to simulate V1

Driving in heavy snow, rain, a dust-storm or other adversarial conditions impacts the ability of the human visual system to recognize objects. Autonomous systems like self-driving vehicles are even more susceptible to such rarely occurring or out-of-distribution input when interacting with the real world. Object recognition is one of the most fundamental problems solved by primates for their everyday functioning. Humans base their decisions on a wide range of bottom-up and top-down cues, ranging from color to texture to an overall ``figure/ground" contour, and on the context that surrounds the object to be recognized \cite{saisan2001dynamic, renninger2004scene, kellokumpu2011recognition, de1998texture, chaaraoui2013silhouette, al2015human, popoola2012video, oliva2007role, zhang2020putting}. Humans combine or seamlessly switch between such cues \cite{mori2004recovering, beleznai2009fast}. These cues help recognize the presence of an ``object," instead of accurately predicting the low-level details about it (e.g. vehicle make, license plate, or text on the rear windshield). The primate visual system is robust to small perturbations in the scene~\cite{zhou2019humans, eidolons} and uses sophisticated strategies to recognize objects with high accuracy and confidence. 

Artificial neural networks (ANNs) learn to recognize objects with only bottom-up cues like contours, color, texture, etc., allowing them to easily exploit ``shortcuts" in the input distribution~\cite{geirhos2018generalisation, geirhos2020shortcut} (eg., a red spherical object is mostly classified as an apple). These shortcuts affect their performance when the objects are distorted by adversarial attacks, limiting their capability to generalize to an out-of-distribution input. ANNs tend to recognize objects both in the presence and in the absence of object structure. This ability also helps them learn from images that appear to be noise to humans \cite{zhou2019humans}.

Brain regions cannot be directly equated to layers in networks~\cite{yamins2016using, dong2018commentary}. While different regions of the brain are primarily responsible for processing different input stimuli~\cite{bear2020neuroscience}, the visual system processes an object as a ``whole" by relying on its contours instead of lower-level features like color. Parts of the objects are assembled, and bounding areas of the key points provide an overall shape for the object~\cite{mumford1, mumford2}. The reverse hierarchy theory states that humans approach visual classification with a holistic approach, looking at ``forest before trees" and then adjust to the lower level details as needed~\cite{hochstein2002view}. Humans calculate the gist of the overall scene before proceeding to do figure-ground segregation and grouping visual objects together~\cite{corbett2023pervasiveness}. Humans also need surprisingly little visual information to classify objects~\cite{ullman2016atoms}. These findings motivate our work --- to see how humans and machines perform when the object structure is altered.

To probe the limits of this gap between human and network performance on object classification, we introduce novel image transformation techniques based on what is known in the psychology literature to affect human vision \cite{edelman1997complex, tarr1998image, grill2001lateral, biederman1991priming, ferrari2007groups}, but which go beyond the currently employed techniques of adversarial attacks in machine vision. Our experiments test the limits to which humans and ANNs can withstand these attacks. We further categorize differences between the strategies employed by both to solve tasks with these transforms. We propose a ranking of these attacks based on humans' ease of solving our recognition task.
% *** asking the question: do networks learn with a structure of an object or lack there of?

\paragraph{Related work }
Ullman et al.~\cite{ullman2016atoms} use ``minimal recognizable images" to test the limits of network performance on object recognition, and show that they are susceptible to even minute perturbations at that level. Rusak et al.~\cite{rusak2020simple} show that an object recognition model that is adversarially trained against locally correlated noise improves performance.  By removing texture information and altering silhouette contours, Baker et al.~\cite{baker2018deep} show that networks focus on local shape features by shuffling the object silhouettes shown to networks and humans. Baradad et al.~\cite{learningtoseebylookingatnoise} try to learn robust visual representations by generating models of noise closer to the distribution of real images. Nguyen et al. \cite{nguyen2015deep} generate images using evolutionary algorithms, and attack the networks pretrained on datasets like Imagenet \cite{imagenet}.

Geirhos et al.~\cite{geirhos2018imagenettrained} found that Imagenet~\cite{imagenet} pretrained ResNets~\cite{resnet} recognize the textures in objects with a high accuracy and minimum attention to the segmentations --- a dog with the texture of elephant is recognized as an elephant by networks, but is recognized as a dog by humans. Scrambled images do not affect the networks very much~\cite{gatys2017texture, brendel2019approximating}, until low-level features are affected~\cite{yu2017sketch, ballester2016performance}. Tolstikhin et al.~\cite{tolstikhin2021mlpmixer} show the use of patches with multi-layer perceptrons can yield performance rivaling Vision Transformers (ViT)~\cite{dosovitskiy2021an}.

Zhou et al.~\cite{zhou2019humans} show that while machines can be fooled by adversarial images, humans tend to use their intuition about objects in classifying images that are ``totally unrecognizable to human eyes". They further hint that these intuitions can be used to guide machine classification. Dapello et al.~\cite{dapello2021neural} show that neural networks with adversarial training and general training routines have geometrical differences in their representations in intermediate layers. In \cite{crowder2022robustness}, authors introduced 5 new transforms with extreme pixel shuffling and found that, barring a few cases, humans perform significantly better than networks on $28 \times 28$ pixel CIFAR100 images. In this work we show that while the trend holds for some transforms on large $320 \times 320$ pixel Imagenette images, there are significant differences in strategies used by humans and machines to recognize objects.

%This is a suggested rewrite for the paragraph
% Networks are susceptible to minute perturbations on the ``minimal recognizable images" level \cite{ullman2016atoms}. Networks also focus on local shape features, found by removing texture information and altering silhouette contours 
% \cite{baker2018deep}. Still, adversarially trained object recognition models against locally correlated noise improves performance \cite{rusak2020simple}.  
% Baradad et al.~\cite{learningtoseebylookingatnoise} try to learn robust visual representations by generating models of noise closer to the distribution of real images.

\vspace{4mm}
\noindent\textbf{Contributions: } Humans and machines use different strategies to recognize objects under extreme image transformations. Humans base their decisions on object boundaries and contours, while networks rely more on low-level features like color and texture.
\begin{itemize}
    % \vspace{-2mm}
    \item We introduce novel image transforms with blocks and image segmentation to simulate extreme adversarial attacks on humans and machines for the task of object recognition.
    % \vspace{-2mm}
    \item We present an extensive study probing the limits of network performance with changes in our transform parameters. We evaluate the performance of ResNet50, ResNet101 and VOneResNet50, as well as 32 human subjects on our transforms.
    % \vspace{-2mm}
    \item We highlight the differences in strategies employed by humans and networks for solving object recognition tasks and present extensive statistical analysis on the performance and confidence of humans and machines on our extreme image transformations.
    \item We propose a ranking for complexity of transforms (and their parameters) as observed by humans and machines, and find that humans recognize objects with contours while machines rely on color/texture, challenging how far network performance is from becoming human-like. 
    % \item We probe the limit of robustness between humans and networks on object recognition tasks through our transforms
    % nobody has done a study at this level with so many variations of transforms with humans
\end{itemize}

% Currently  many neural networks are being built in a way that tries to mirror the function of the primate visual cortex, in order to get a better understanding of the functionalities. So we wanted to see if there were correlations between how a model identifies and object and how a human does. Ullman found that when minute changes are made to an image the ability for models to identify what the object is drops significantly, while humans are still able to identify what that object is. Thus, coming to the conclusion that the current models are not able to recognize images at the scale of a human due to models losing the identifying features they use to differentiate between objects when minimal images are used{*** cite atoms of rec ***}. Bowers et al. also explains that although DNNs are most accurate in classifying images and predicting human error, having high performance on brain and behavioral benchmarks may share on little intersection with the function of biological vision{*** cite deep problems with nn models of human vision ***}. However, VOneNet was designed based off of area v1 in the brain{}. This model has been claimed to have similar functions as area V1 in the brain. 

%------------------------------------------------------------------------
\section{Extreme Image Transformations}
\label{sec:transformations}

% *** Talk a bit about related work, and motivate why such extreme transformations are needed. Can build from Atoms of Recognition min recognizable configurations (MIRC).  - DONE
A recent trend in deep networks is to meet or exceed the performance of humans on a given task \cite{he2015delving, taigman2014deepface, mnih2015human}. Papers in the past decades have claimed in-silico implementations of primate visual cortex \cite{douglas1991functional, wilson1991computer, bednar2012building}. Ullman et al.~\cite{ullman2016atoms}, however, found that minute changes in the images can significantly impact network performance, while having little or no influence on humans. They showed that human performance remains almost untouched at the scale of minimum recognizable configurations (MIRC). This behaviour could be due to networks' dependence on background and other extra features that they learn to solve tasks. Humans base their decisions on complete and partial presence of features at different scales~\cite{wang2008perceptual, georgeson2007filters, witkin1987scale, lindeberg2013scale, ekstrom2017human}. For example, a silhouette of a zebra can be classified as a horse, but a close look at the ears (not even entire face) might be enough to tell the difference. 

Our work asks if humans and machines show a similar response on an object recognition task, without physically breaking down the images into smaller independent images with the atomic representation of the object class (as done in~\cite{ullman2016atoms}). We further probe the images at different scales within the blocks and segments by varying block size, the probability of shuffling a pixel, and by interchanging the complete regions with each other. 

\begin{figure}
  \centering \includegraphics[width=1.\textwidth]{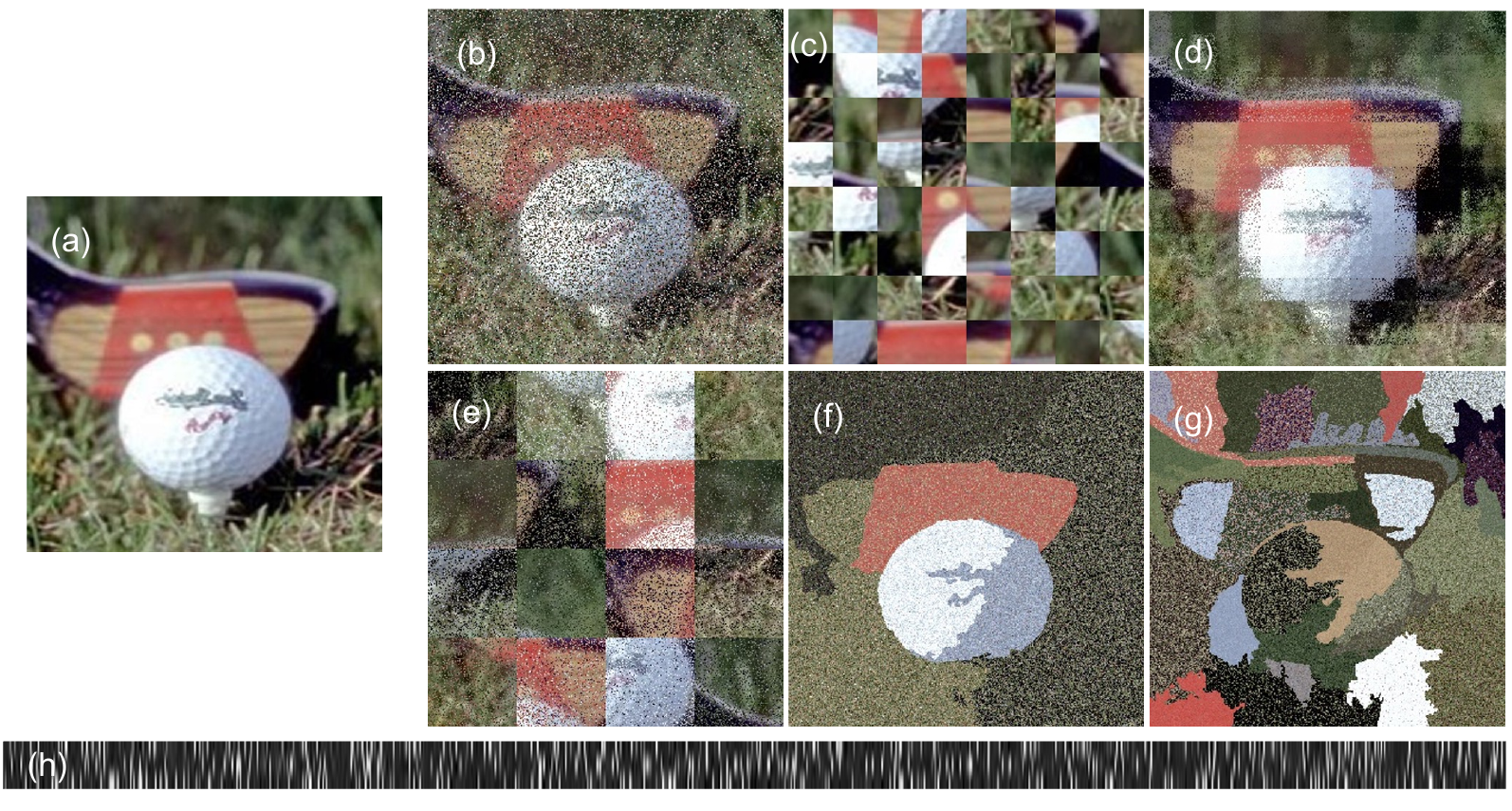}
  \caption{Extreme Image Transformations applied to an Imagenette image of category \textit{Golf Ball}. (a) non-transformed baseline image, (b) Full Random Shuffle with probability 0.5, (c) Grid Shuffle with grid size 40x40, (d) Within Grid Shuffle with block size 40x40 and probability 0.5, (e) Local Structure Shuffle with block size 80x80 and probability 0.5, (f) Segmentation Within Shuffle with 16 segments and probability 1.0, (g) Segmentation Displacement Shuffle with 64 segments, (h) Color Flatten.}
  \label{fig:transforms_all}
\end{figure}

We introduce seven novel image transformations (Figure \ref{fig:transforms_all}), to test the limits of human and machine vision on the object recognition task with distorted image structures. Our transforms can be controlled by three independent variables -- i) \textbf{Block size} (or number of segments in case of segmentation shuffles that are described below), ii) \textbf{Probability} of individual pixel shuffle and iii) \textbf{Moving} blocks or regions to another location or not. Traversing this 3-dimensional space leads to a wide variety of variations in the visual perception of objects for humans and machines (Figure \ref{fig:rankingsurface}). Our transformations can be split into block and segmentation shuffle:

\texttt{block\_shuffle(block\_size [\#pixels], pix\_shuffle\_prob [0..1], block\_shuffle [0/1])}, and 

\texttt{segment\_shuffle(segments [\#partitions], pix\_shuffle\_prob [0..1], region\_shuffle [0/1])}

% should be shown as a diagram on multiple axes?
\noindent\textbf{Full Random Shuffle} moves pixels within the image based on a specified probability (range: [0.0 - 1.0]), disregarding any underlying structural properties of the image. For a shuffle probability of 0.5, each pixel’s location has a 50\% chance of being shuffled, while a shuffle probability of 1.0 moves every pixel around, with the image looking like random noise. Lower probability alters local structure while higher probability alters the global structure of the image. Figure \ref{fig:transforms_all}(b).

\noindent\textbf{Grid Shuffle} divides the image into blocks of equally sized squares. The divided units are shuffled and rearranged to create an image of the same size as input. Block length is chosen out of [20, 40, 80, 160] pixels. Grid Shuffle alters only the global structure of the image. Figure \ref{fig:transforms_all}(c).

\noindent\textbf{Within Grid Shuffle} divides the image into blocks (similar to Grid Shuffle), but does not shuffle the blocks. Instead, it shuffles the pixels within the blocks with a specified probability. Pixel shuffling within the block is similar to Full Random Shuffle, considering each unit in the block to be an individual image. Grid size and probability of shuffle is in [20, 40, 80, 160] pixels and [0.0 - 1.0] respectively. Alters only the local structure of the image. Figure \ref{fig:transforms_all}(d).

\noindent\textbf{Local Structure Shuffle} is a combination of Within Shuffle and Grid Shuffle. It divides the image into blocks (like Grid Shuffle), shuffles the pixels within the blocks (like Full Random Shuffle), and further shuffles the positions of the blocks. Alters both global and local structure of the image. Figure \ref{fig:transforms_all}(e).

\noindent\textbf{Color Flatten} separates the three RGB channels of the image and flattens the image pixels from 2-dimensional \text{$N \times N$} to three channel separated \text{1-dimensional} vectors of length \text{$N*N$} in row-major order. Alters both global and local structure of the image. Figure \ref{fig:transforms_all}(h).

\noindent\textbf{Segmentation Within Shuffle} builds on the grid shuffle paradigm by segmenting the image into regions based on superpixels \cite{achanta2010slic, achanta2012slic}. The pixels within the region are shuffled with a specified probability in the range [0.0 - 1.0]. The number of segments is picked from [8, 16, 64]. Figure \ref{fig:transforms_all}(f).

\noindent\textbf{Segmentation Displacement Shuffle} segments the image into regions (8, 16 or 64) based on superpixels \cite{achanta2010slic, achanta2012slic}. The pixels within each region are shuffled and placed into other regions. The number of pixels in every region can differ significantly prohibiting a smooth displacement when moving pixels from smaller region to a larger region. We solve this problem by re-sampling a number of pixels equal to the difference in number of pixels between larger and smaller region again from the smaller region. We also shuffle them with all the pixels from the smaller region and arrange them in the larger region. Moving from larger to smaller region, we drop the extra pixels. Figure \ref{fig:transforms_all}(g).

\paragraph{Local vs global manipulations} It is difficult to precisely categorize transforms into local or global manipulators. Our approach holds that local transforms manipulate the low level features of the image (not necessarily only texture, but also some borders of the object), while global manipulations alter the overall shape of the object. For example, a Full Random Shuffle with a low probability of say 0.3 can be broadly categorized as a local manipulator, but with a probability of 1.0, the same transformation changes the global structure. Humans are known to easily switch between local and global structures when performing object recognition, while networks generally do not have a way of doing that. To this end, we also rank our transforms based on human accuracy, which favors preservation of global structure, whereas the networks' rankings tend to rely more on local structure.

\section{Model Selection}
\label{sec:models}
We tested ResNet50, ResNet101~\cite{resnet}, and VOneResNet50~\cite{vonenet} for our experiments with baseline (no shuffle) and transformed images.
VOneResNet50 was selected for its claims of increasing robustness of Convolutional Neural Network (CNN) backbones to adversarial attacks by preprocessing the inputs with a VOneBlock -- mathematically parameterized Gabor filter bank -- inspired by the Linear-Nonlinear-Poisson model of V1. It also had a better V1 explained variance on the Brain-Score~\cite{brainscore} benchmark at the time of our experiments. 
We chose ResNet50 since it formed the CNN backbone of VOneResNet50 and we wanted to test the contribution of non-V1-optimized part of VOneResNet50.
Subsequently we chose ResNet101 for its high average score on brain-score in terms of popular off-the-shelf models that are widely in use, and for its larger capacity compared to ResNet50.

%------------------------------------------------------------------------

\section{Experiments}
\label{sec:experiments}
%setup/dataset
% why cifar, what experiments

\noindent\textbf{Setup: } We evaluated ResNet50, ResNet101, and VOneResNet50 on the Imagenette dataset \cite{imagenettegit} against baseline images (without transforms), six shuffle transforms, and Color Flatten transform described in \textsection\ref{sec:transformations}. Imagenette is a subset of 10 unrelated Imagenet \cite{imagenet} classes. We used the default train-test split of 9469 training images and 3925 test images from the dataset, distributed over 10 classes. Each image has 3 channels and $320 \times 320$ pixels. We did \textit{not} separate a validation set for network fine-tuning to mimic how humans only see a small subset of objects and then recognize them in the wild, without fine-tuning their internal representations. To further this claim, we also performed 0-shot experiments with Imagenet pretrained networks on images processed with our transforms (please see \textsection\ref{app:saliency}). We used Imagenette for training and testing, given the models we selected are all trained on the larger Imagenet dataset. 

% justify the use of CIFAR100 as dataset

%training
\noindent\textbf{Training: } We trained each model on the baseline and transformed images using the default hyperparameters listed in the PyTorch repositories of the respective models. All models were trained for 70 epochs, with a learning rate of 0.1, momentum set to 0.9 and weight decay of $10^{-4}$.  For the Grid Shuffle transformation, we used four block sizes -- $20 \times 20$, $40 \times 40$, $80 \times 80$ and $160 \times 160$ (dividing image into 4 blocks). For Within Grid Shuffle and Local Structure Shuffle we used a combination of four block sizes ($20 \times 20$, $40 \times 40$, $80 \times 80$ and $160 \times 160$) with a shuffle probability of 0.5 and 1.0 for each block. For Full Random Shuffle we used shuffle probabilities of 0.5, 0.8 and 1.0. We used 23 different block transformations. 

%For all transforms involving a grid, we limited ourselves to perfect squares given the square shape of the image and ease of handling with humans and networks. *** limitations

%histogram
In the Color Flatten transform, we separated the image channels and flattened the 2D array to 1D in row-major order. We added an additional Conv1D input layer to the networks to process the 1D data.

Humans can base object recognition decisions on the boundaries of objects \cite{edelman1997complex, tarr1998image, biederman1991priming, ferrari2007groups,hubel1962receptive, hubel1963shape, wiesel1963single, hubel1963receptive, wiesel1963effects, tanaka1997mechanisms, grill2001lateral}. We wanted to test networks in the same settings by using superpixels to segment objects into varying number of regions and shuffling pixels within/across regions. We trained and tested each of the three models on the segmentation shuffles described in Section \ref{sec:transformations}. For Segmentation Displacement Shuffle we segmented the images into 8, 16 and 64 regions. For Segmentation Within Shuffle, we used a combination of 8, 16 and 64 regions, with a pixel shuffle probability of 0.5 and 1.0. We trained and tested on 9 unique segmentation transformations.
All networks were trained end-to-end using only the respective transform (and its hyperparameters), without sharing any hyperparameters across same or different types of transforms.

%human trial -- to be done, describe the process of trials -- warmup, test; how the system was setup, etc. Put extra details to supplementary

\section{Human Study}
\label{sec:humanexperiments}
To investigate mechanisms used by humans for solving object recognition task under adversarial attacks and compare it to networks, we ran a psychophysics study with 32 participants on a Cloud Research's Connect platform. We randomly sampled 3 images from each of the transform-parameter pair to test the subjects, after training them on 11 sample images from Imagenette dataset. We used the same $320 \times 320$ pixel resolution images for both humans and networks, and presented all subjects with the same 10 classes to choose from. We also asked the subjects to indicate their confidence about their response on a scale of 1 to 5. The classes were randomly shuffled on every trial. We gave feedback to subjects after every trial during training phase, but not during the testing phase. We timed their responses on test trials, but they were asked to complete the trials at their own pace. We turned off the timer after every 10 test trials to allow for breaks if needed. We used the same set of three unique images to use with a particular transformation-parameter pair to show to all participants. This means the participants saw the same set of 102 unique images during the entire study. None of the images were repeated during the training or testing phase to avoid learning any kind of biases in object structures for that exact image. For more details about experiment setup, participant filtering and statistical tests, see \textsection\ref{sec:human_setup} and \textsection\ref{supsec:statanalysis}. 
\section{Results}
\begin{table}[ht]
\caption{Test accuracy for models trained on Imagenette dataset with Block transforms}
\label{table:testacc_block}
\begin{tabular}{lccccc}
\toprule

\multirow{2}{*}{\textbf{Transform}}  & \multirow{2}{*}{\textbf{P}} & \multirow{2}{2em}{\textbf{Grid Size}} & \multicolumn{3}{c}{\textbf{Accuracy (in \%)}}\\  \cmidrule{4-6}
& & &\textbf{ResNet50} & \textbf{ResNet101} & \textbf{VOne}  \\  
\midrule

                            %   p       size        50          101         vone
\text{Baseline}             &           &           & 86.24     & 85.61     & 72.2      \\ \cmidrule{1-6}
\text{Full Random Shuffle}       & 0.5       &           & 83.9      & 83.44     & 72.22     \\ 
                            & 0.8       &           & 60.66     & 62.54     & 49.78     \\ 
                            & 1.0         &           & 47.46     & 49.3      & 31.62     \\ \cmidrule{1-6}
\text{Grid Shuffle}         &           & 20x20     & 84.99     & 84.46     & 49.47     \\
                            &           & 40x40     & 86.78     & 86.29     & 52.28     \\ 
                            &           & 80x80     & 86.29     & 86.37     & 70.19     \\ 
                            &           & 160x160   & 85.61     & 85.2      & 75.06     \\ \cmidrule{1-6}
\text{Within Grid Shuffle}  & 0.5        & 20x20     & 83.67     & 83.69     & 70.93     \\
                            &           & 40x40     & 83.11     & 82.9      & 72.25     \\ 
                            &           & 80x80     & 83.11     & 82.9      & 74.78     \\ 
                            &           & 160x160   & 83.44     & 83.16     & 70.78     \\ \cmidrule{2-6} 
                            & 1.0         & 20x20     & 79.11     & 78.57     & 69.45     \\
                            &           & 40x40     & 72.51     & 71.92     & 67.64     \\ 
                            &           & 80x80     & 65.58     & 64.2      & 56.31     \\ 
                            &           & 160x160   & 53.89     & 54.08     & 40.61     \\ \cmidrule{1-6}
\text{Local Structure Shuffle}   & 0.5        & 20x20     & 76.89     & 75.95     & 58.44     \\
                            &           & 40x40     & 81.78     & 80.74     & 57.99     \\ 
                            &           & 80x80     & 82.55     & 82.11     & 65.55     \\ 
                            &           & 160x160   & 80.66     & 81.15     & 60.46     \\ \cmidrule{2-6} 
                            & 1.0         & 20x20     & 68.18     & 68.19     & 49.4      \\
                            &           & 40x40     & 66.24     & 65.22     & 45.27     \\ 
                            &           & 80x80     & 58.11     & 57.5      & 44.43     \\ 
                            &           & 160x160   & 51.36     & 49.53     & 36.0      \\ \cmidrule{1-6}
\text{Color Flatten}        &           &           & 75.67     & 73.83     & 59.03     \\ 
\bottomrule

\end{tabular}
\end{table}

\begin{table}[ht]
\caption{Test accuracy for models trained on Imagenette dataset with Segmentation transforms}
\label{table:testacc_seg}
\begin{tabular}{lcp{13mm}ccc}
\toprule
% todo fix segments, formatting looks weird
\multirow{2}{*}{\textbf{Transform}}  & \multirow{2}{*}{\textbf{P}} & \multirow{2}{2em}{\textbf{Segments}} & \multicolumn{3}{c}{\textbf{Accuracy (in \%)}}\\  \cmidrule{4-6}
& & &\textbf{ResNet50} & \textbf{ResNet101} & \textbf{VOne}  \\  
\midrule

                            %   p       size        50          101         vone
Segmentation Displacement Shuffle &           & 8        & 42.98     & 38.39     & 4.54     \\
                             &           & 16       & 45.53     & 45.65     & 4.01     \\ 
                             &           & 64       & 53.68     & 53.89     & 19.97    \\ \cmidrule{1-6}
Segmentation Within Shuffle       & 0.5       & 8        & 71.17     & 72.79     & 4.63     \\
                             &           & 16       & 72.95     & 72.95     & 4.73     \\ 
                             &           & 64       & 75.51     & 76.07     & 4.58     \\ \cmidrule{2-6} 
                             & 1.0       & 8        & 51.93     & 48.41     & 4.91     \\
                             &           & 16       & 57.99     & 48.41     & 4.91     \\ 
                             &           & 64       & 67.94     & 67.86     & 4.75     \\ 
\bottomrule
\end{tabular}
\end{table}

\begin{figure}
  \centering \includegraphics[width=.49\textwidth]{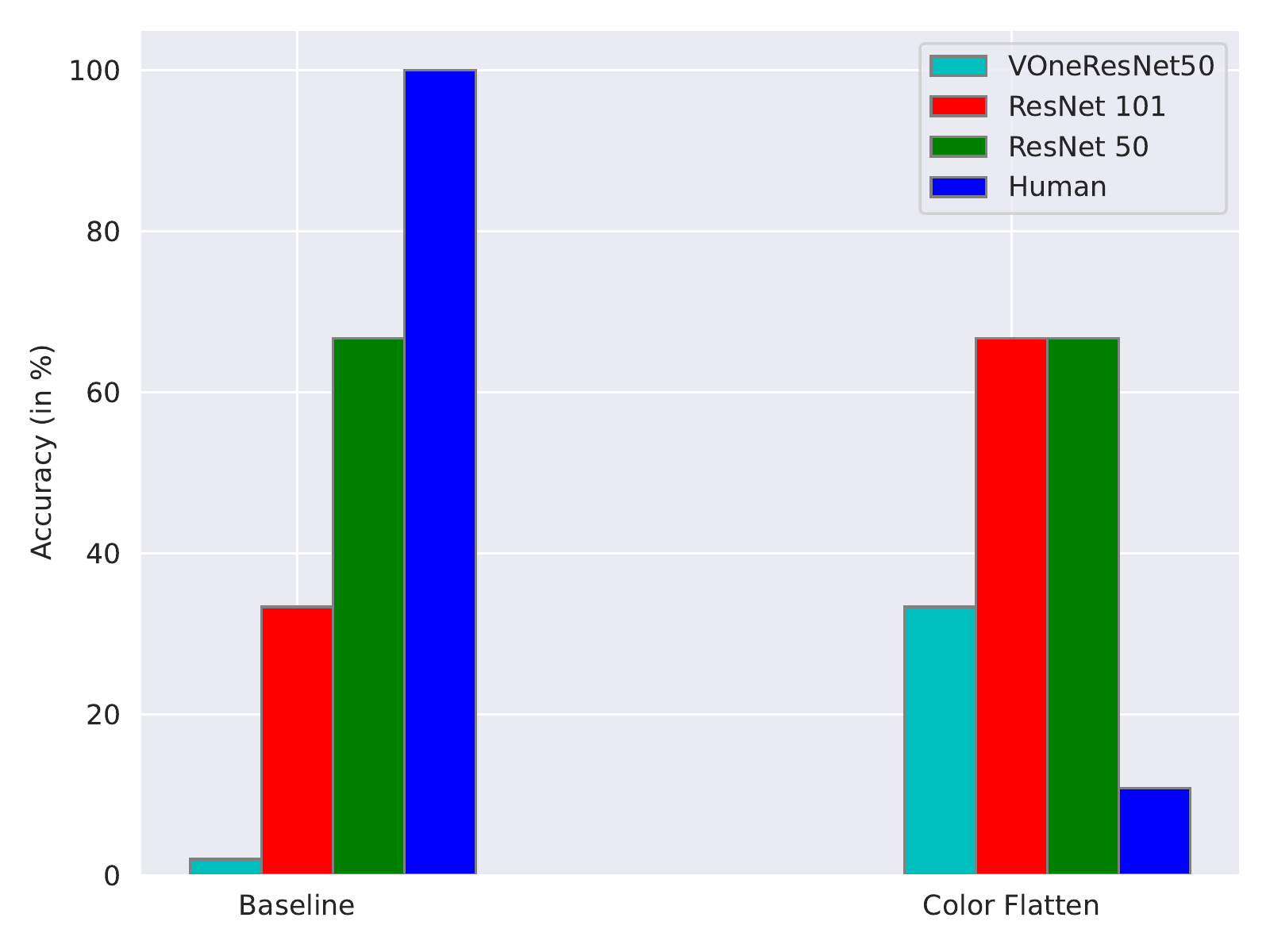}
    \centering \includegraphics[width=.49\textwidth]{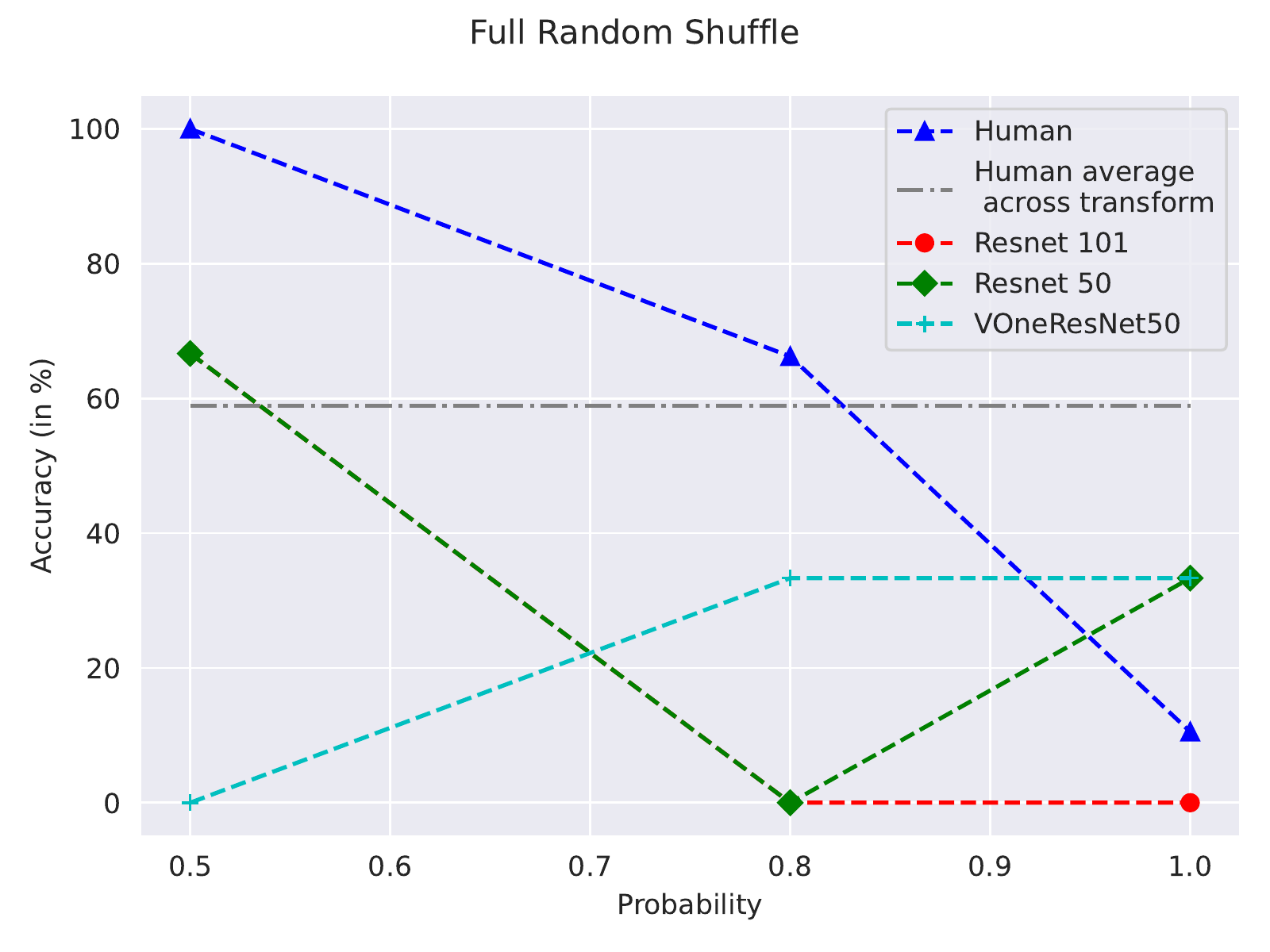}

  \caption{Performance of humans and networks on same images for Baseline and Color Flatten transforms (left) and Full Random Shuffle transform as a function of probability (right). See table \ref{tab:humanresponseblocktransforms} and text for details.} 
  \label{fig:baseline_colorflattern_fullrandomshuffle}
\end{figure}

\begin{figure}
  \centering \includegraphics[width=.8\textwidth]{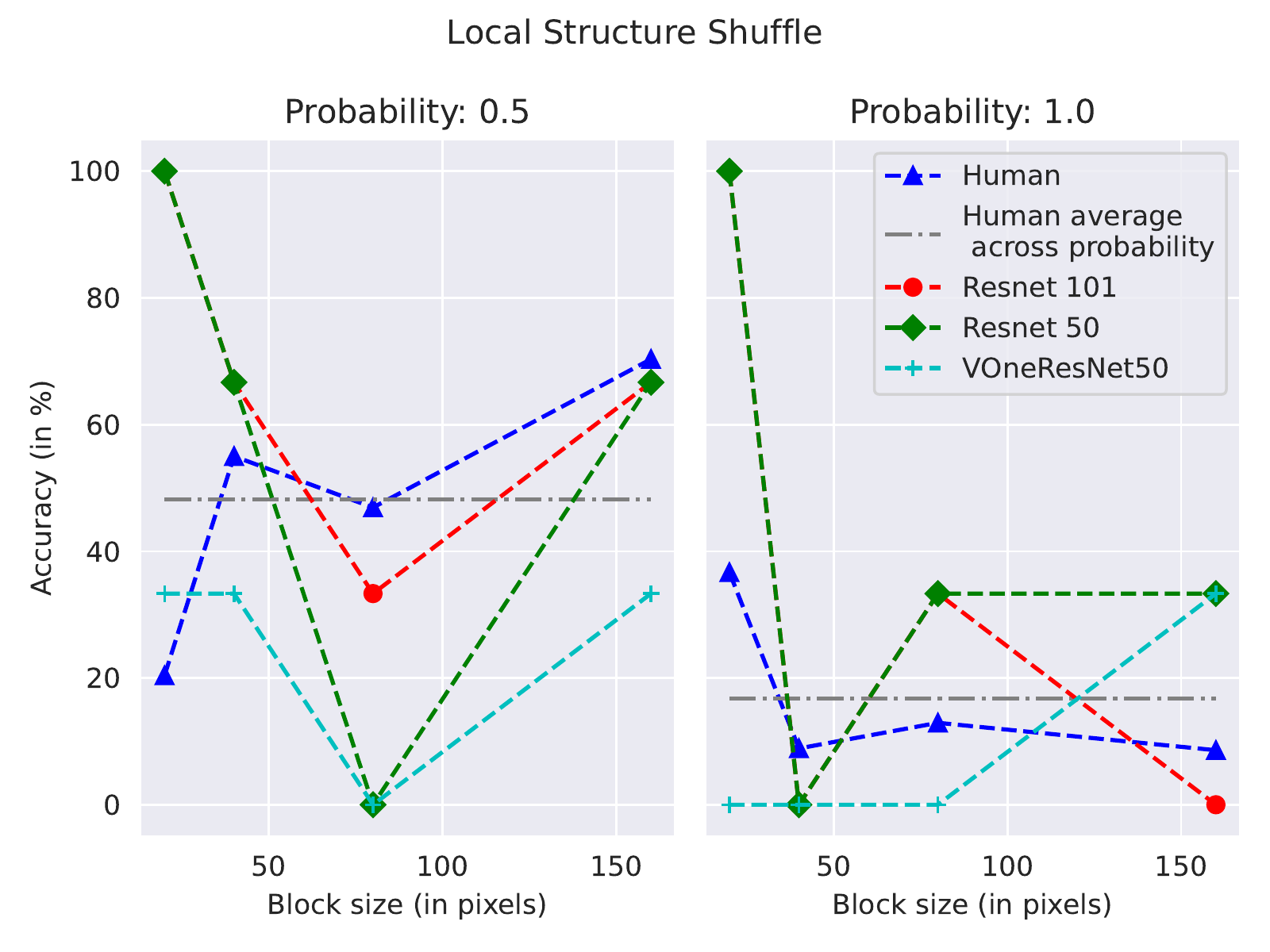}
  \caption{Performance of humans and networks on same images for Local Structure Shuffle with probability 0.5 (left) and probability 1.0 (right) as a function of block size. See table \ref{tab:humanresponseblocktransforms} and text for details.}
  \label{fig:localstructure}
\end{figure}

\begin{figure}
  \centering \includegraphics[width=.49\textwidth]{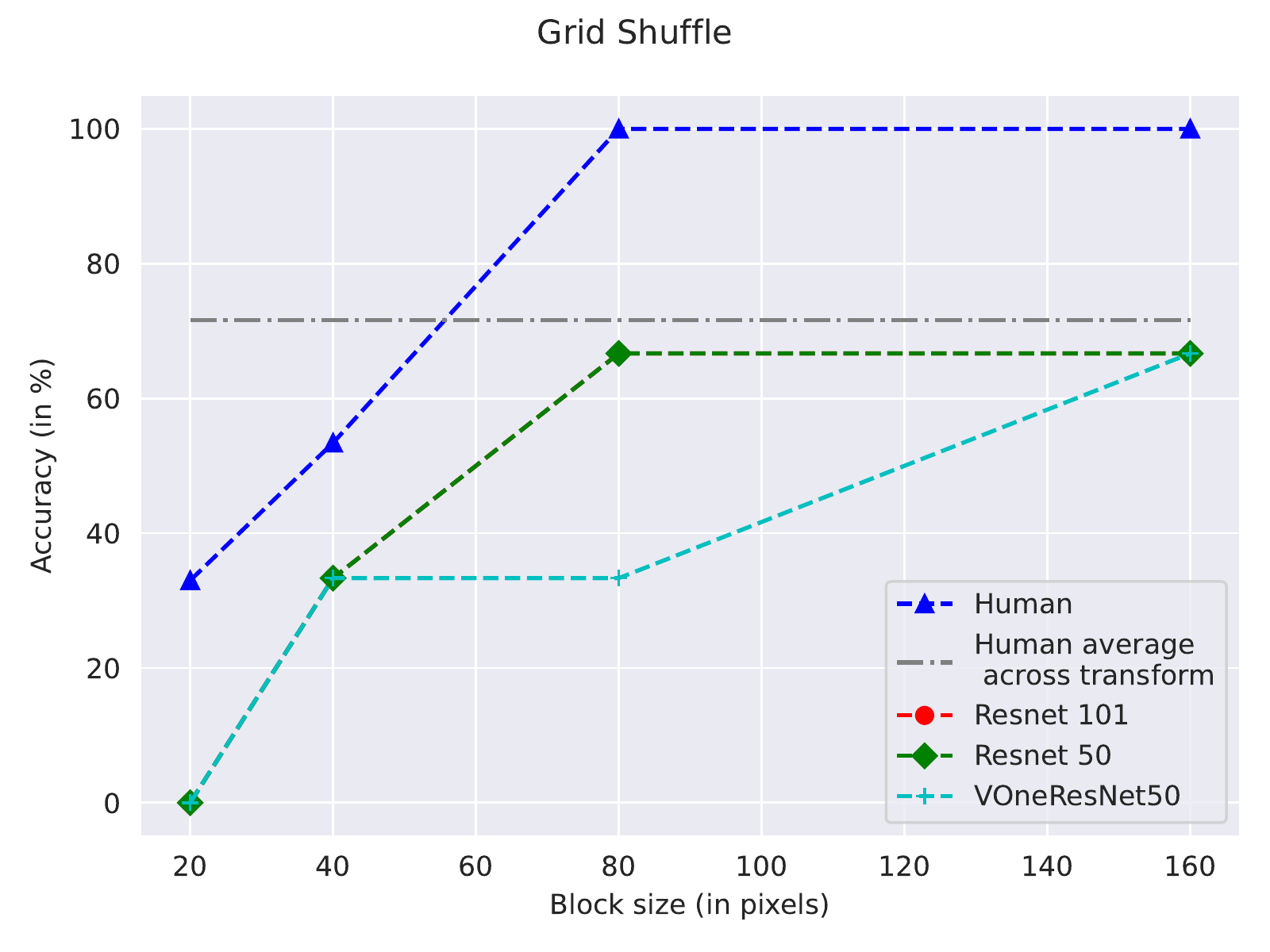}
    \centering \includegraphics[width=.49\textwidth]{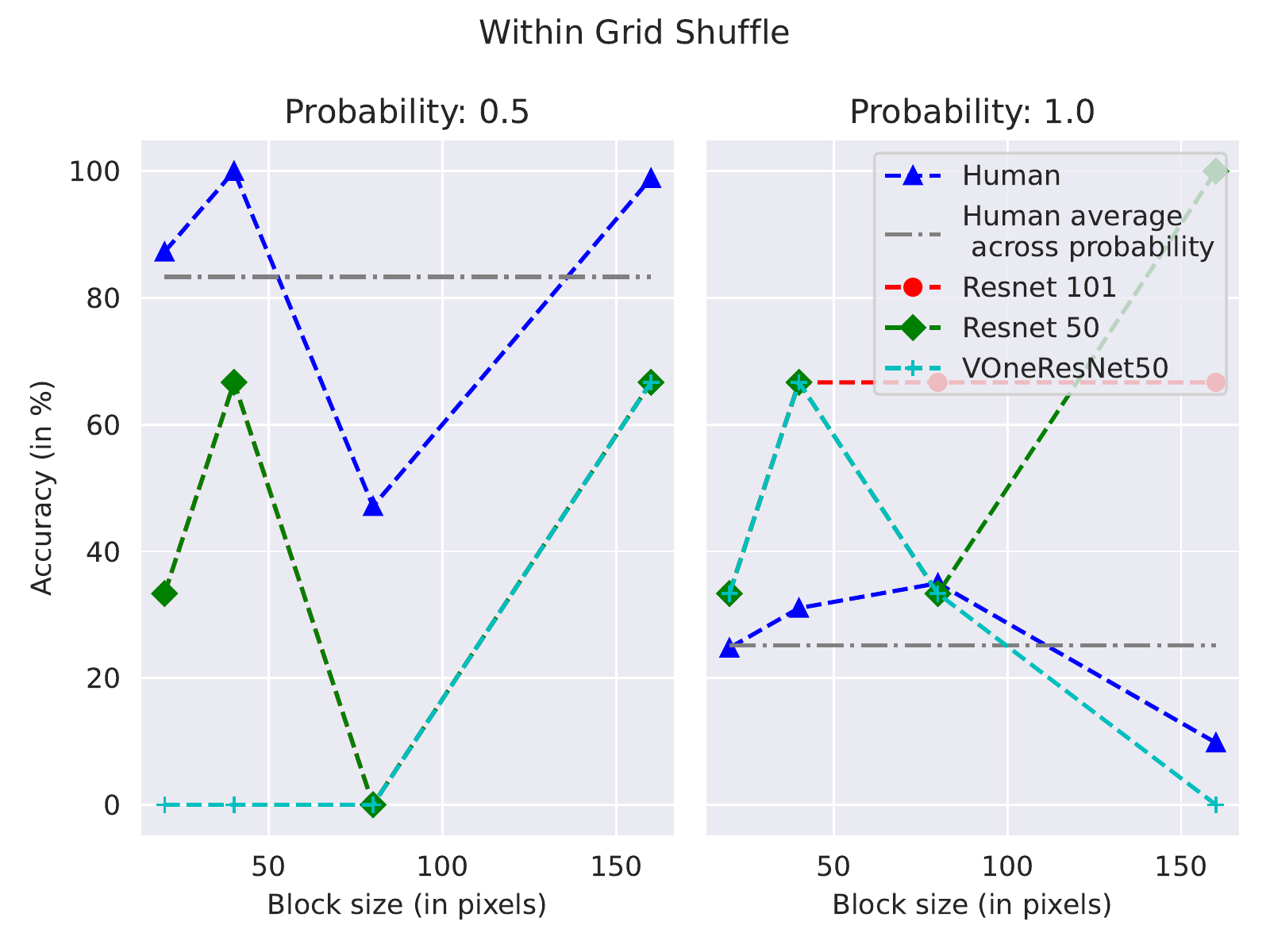}

  \caption{Performance of humans and networks on same images on Grid Shuffle as a function of block size (left); and Within Grid Shuffle (right). The plot for Within Grid Shuffle shows performance on both probabilities as a function of block size. See table \ref{tab:humanresponseblocktransforms} and text for details.}
  \label{fig:gridwithingrid}
\end{figure}

\begin{figure}
  \centering \includegraphics[width=.49\textwidth]{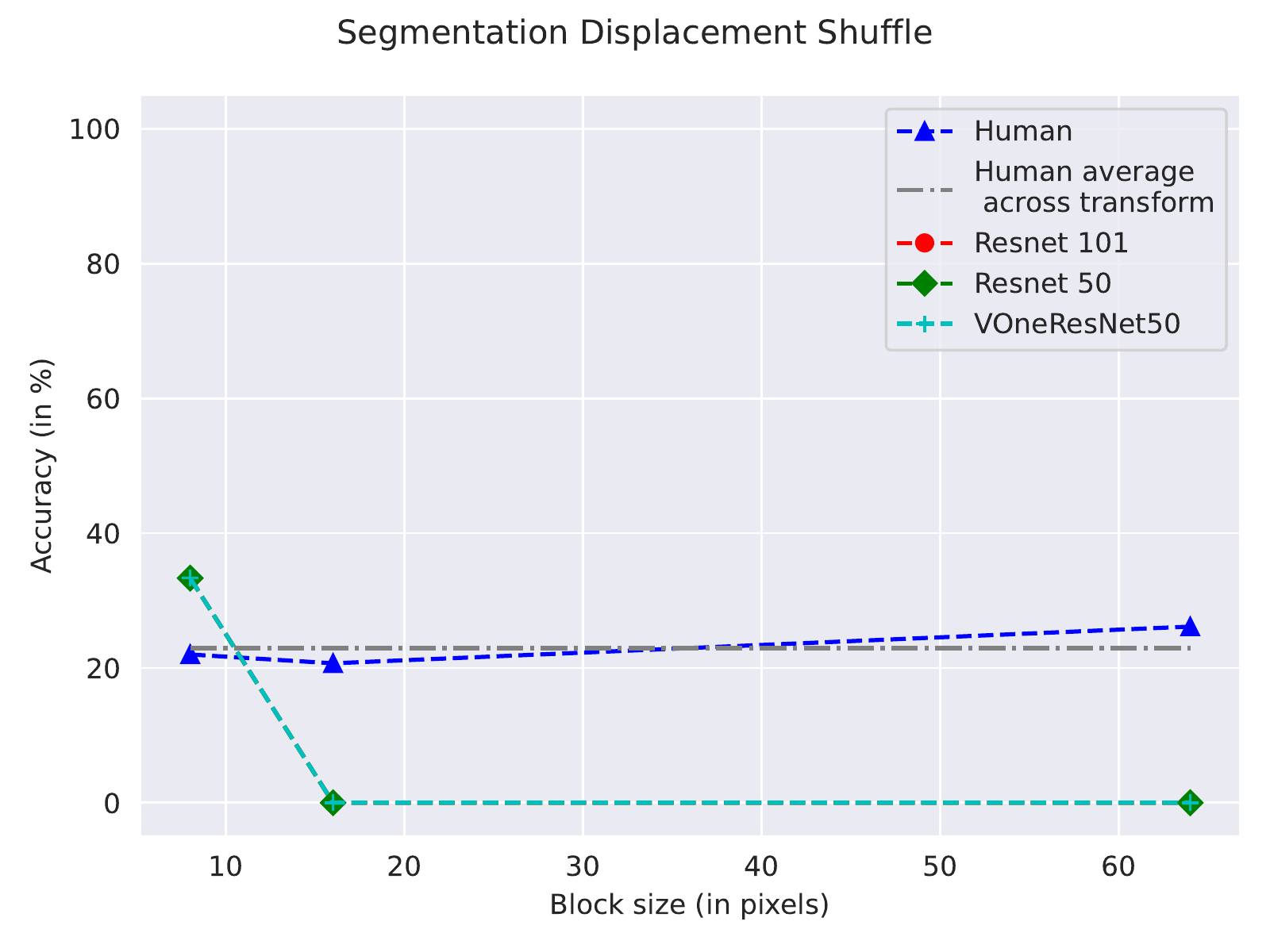}
    \centering \includegraphics[width=.49\textwidth]{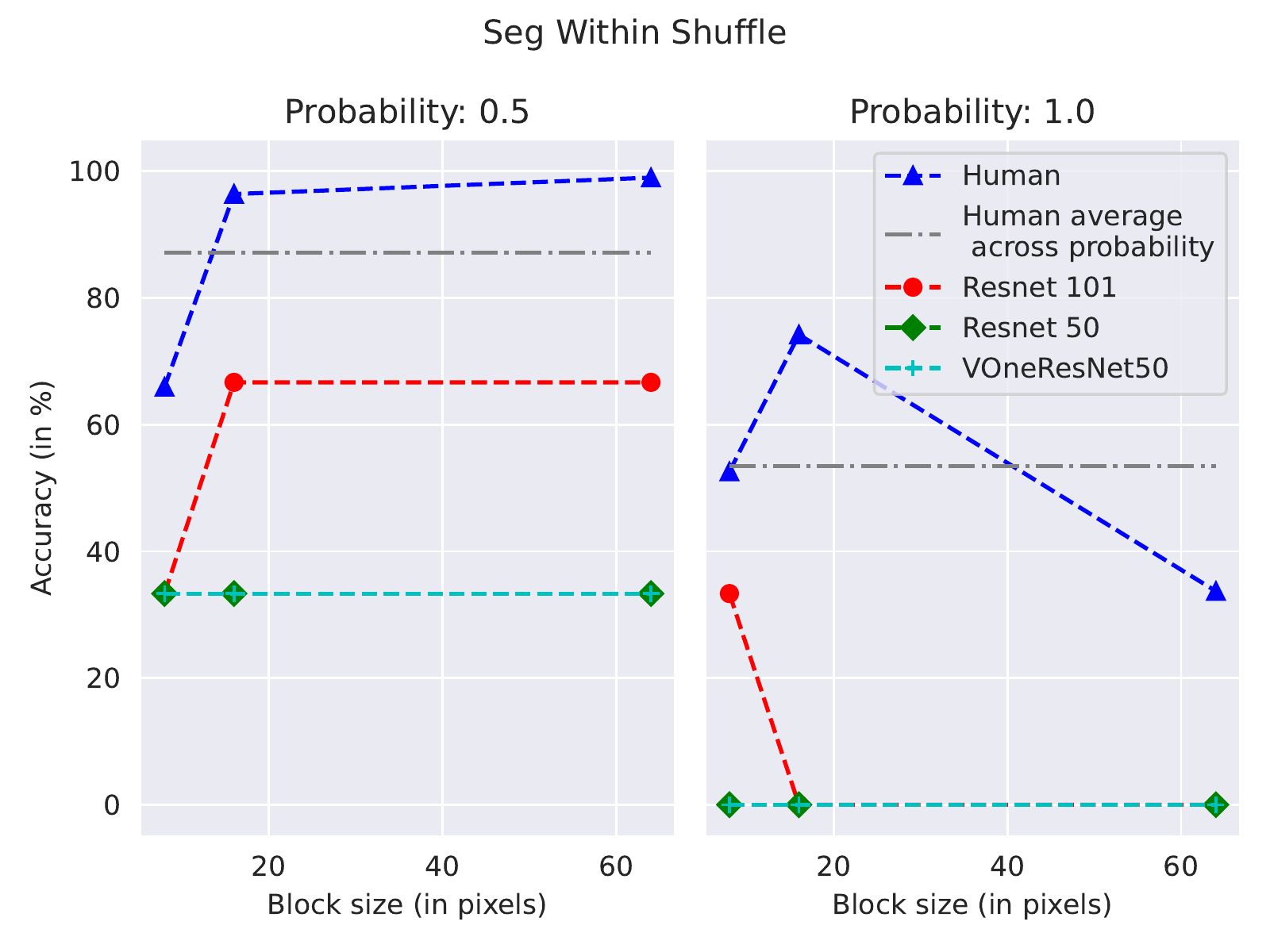}

  \caption{Performance of humans and networks on same images with Segmentation transforms. Segmentation Displacement Shuffle is presented as a function of block size (left); and Segmentation Within Shuffle (right). The plot for Segmentation Within Shuffle shows performance on both probabilities as a function of block size. See table \ref{tab:humanresponsesegmentationtransforms} and text for details.}
  \label{fig:segmentationshuffleboth}
\end{figure}

ResNet50 performs the best on baseline (without transforms) Imagenette test images, with 86.2\% accuracy, followed by ResNet101 and VOneResNet50 (Table \ref{table:testacc_block}). The trend is constant across all transforms for VOneResNet50, wherein ResNets on an average perform about 25\% better than VOneResNet50. Transformations start reducing the performance of networks. For Full Random Shuffle, the performance decreases by only about 2\% with a 0.5 shuffle probability (equal chance of every pixel either moving or staying in the same location), implying that the signal to noise ratio might still be the same as the original image. Increasing the shuffle probability to 0.8 and to 1.0 affects the performance most, reducing to almost half of the original performance. ResNets perform significantly better VOneResNet50. For Grid Shuffle, ResNets stay constant and at par with their baseline performance across all block sizes, while VOneResNet50 suffers from a decrease in block sizes. In case of Color Flatten, our most extreme structure destroying transformation, the performance drops by about 11\% for each network compared to their baselines. The networks still perform above chance, implying that recognition is handled independently of the object's structure.

Changing probability and block sizes together, we find that Local Structure Shuffle is affected more than the Within Grid Shuffle. In case of Within Grid Shuffle, only local structure is altered inside the blocks. The performance trend is reversed compared to the Grid Shuffle, such that an increase in block size reduces performance for shuffle probability of 1.0, but stays constant for a shuffle probability of 0.5. The Local Structure Shuffle alters both the local and global structure of the object. For a shuffle probability of 0.5, the performance seems to be increasing with an increase in block size, given larger block sizes help keep more pixels together during convolutional operations, while a probability of 1.0 reverses that trend, reducing the accuracy with an increase in block size.

Following our experiments about fixed block sizes, we wanted to probe the networks with representations that primates are more comfortable with during object recognition \cite{edelman1997complex, tarr1998image, grill2001lateral, biederman1991priming, ferrari2007groups, hubel1962receptive, hubel1963shape, wiesel1963single, hubel1963receptive, wiesel1963effects}. We repeated similar experiments with our segmentation transforms (Table \ref{table:testacc_seg}). Interestingly, VOneResNet50 suffers the most by this change, dropping the accuracy by over 60\% to single digits. For our Segmentation Displacement Shuffle, we found the networks showed an improved performance with a decrease in the size of segments, again implying better performance despite higher structure alterations locally. We observe a similar trend in case of Segmentation Within Shuffle. ResNets show a greater accuracy in this case compared to Segmentation Displacement Shuffle, but with a similar decrease in performance with an increase in shuffle probability. (Please see \textsection\ref{app:saliency} for saliency maps of Imagenet pretrained networks on our transforms.)

\begin{wrapfigure}{r}{0.5\textwidth}
  \begin{center}
    \includegraphics[width=0.48\textwidth]{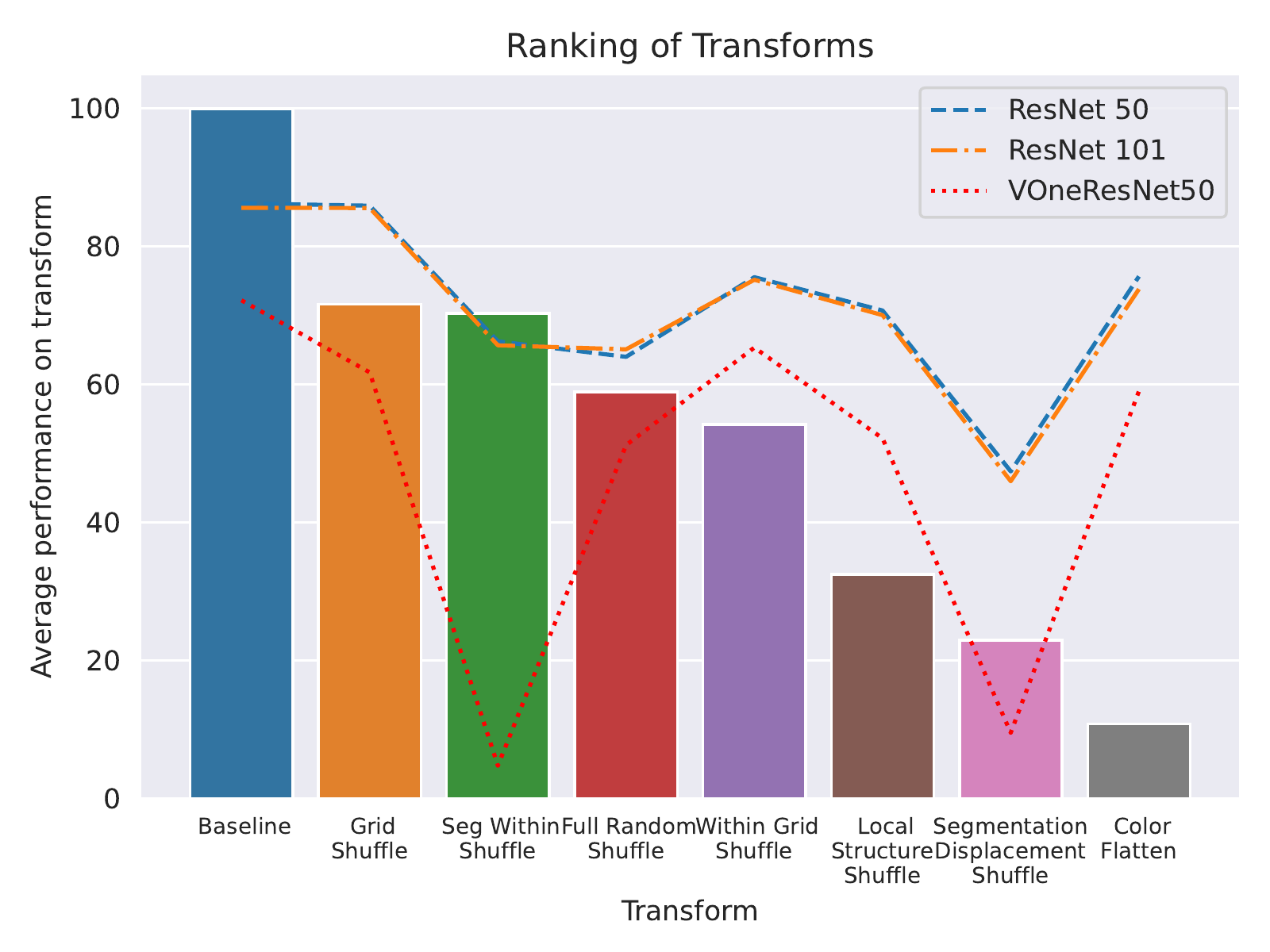}
  \end{center}
  \caption{Average performance per transform for both humans and networks, sorted by human performance in descending order. Average performance for humans is shown as bars while that for networks is overlaid as lines. None of the networks agree with humans on a transform ranking. Individual rankings in Table \ref{suptab:transformranking_coarse}}
  \vspace{-2mm}
  \label{fig:transform_ranking_coarse}
\vspace{-4mm}
\end{wrapfigure}

\noindent\textbf{Comparison with Human responses } Human subjects show no correlation with the networks' performance (Tables \ref{tab:humanresponseblocktransforms}, \ref{tab:humanresponsesegmentationtransforms} and \ref{table:ttestcorrresponses}). The trends in performance are asymmetrical between the two (Figure \ref{fig:transform_ranking_coarse}). Humans perform with a perfect score on baselines and Full Random Shuffle with 0.5 shuffle probability. Human accuracy declines, but is better than networks on a 0.8 shuffle probability case, while it is random at best with a 1.0 shuffle probability (Figure \ref{fig:baseline_colorflattern_fullrandomshuffle}). 

On Grid Shuffle, humans show an increase in performance with an increase in block sizes, reaching a perfect accuracy at block sizes 80 and above, a trend similar to networks but at differing accuracies. On Within Grid shuffle with 0.5 shuffle probability, the accuracy only dips for a block size of 80, but remains better than networks otherwise (the networks have a constant performance). With a shuffle probability of 1.0, the performance is much lower than the networks, with a non-monotonic trend (Figure \ref{fig:gridwithingrid}). 

For Local Structure Shuffle with 0.5 shuffle probability, we see a non-monotonic trend with a much lower performance compared to the networks (Figure \ref{fig:localstructure}). The trend remains similar for the 1.0 shuffle probability case, with numbers comparable to Full Random Shuffle 1.0 probability. Color Flatten also affects the human perception to the level of random decision (Figure \ref{fig:baseline_colorflattern_fullrandomshuffle}). 

For our segmentation displacement cases we see that humans consistently perform better than networks,  indicative of the human visual system's reliance on contours for object recognition. When displacing the shuffled pixels across regions, we see human accuracy plummeting to lower than ResNets. When only shuffling within the regions, human performance is very close to the perfect score in the 0.5 shuffle probability case, but takes a hit with the 1.0 shuffle probability case (Figure \ref{fig:segmentationshuffleboth}). The performance in all cases is much higher than that of VOneResNet50 -- the network claiming to explain V1 variance.

\noindent\textbf{How different are strategies used for object recognition by humans and machines? } Humans show a higher performance on certain images compared to machines, while machines show near baseline performance on images that can be classified as noise at best by humans. To answer the question about the strategies employed by humans and machines to solve object recognition task, we evaluated both humans and machines on the same set of images. We additionally asked how confident they were with their decision of selecting the object class present in the image. As expected, the human confidence scores plummeted with an increase in complexity of the transform. 

\begin{wrapfigure}{r}{0.5\textwidth}
  \begin{center}
    \includegraphics[width=0.48\textwidth]{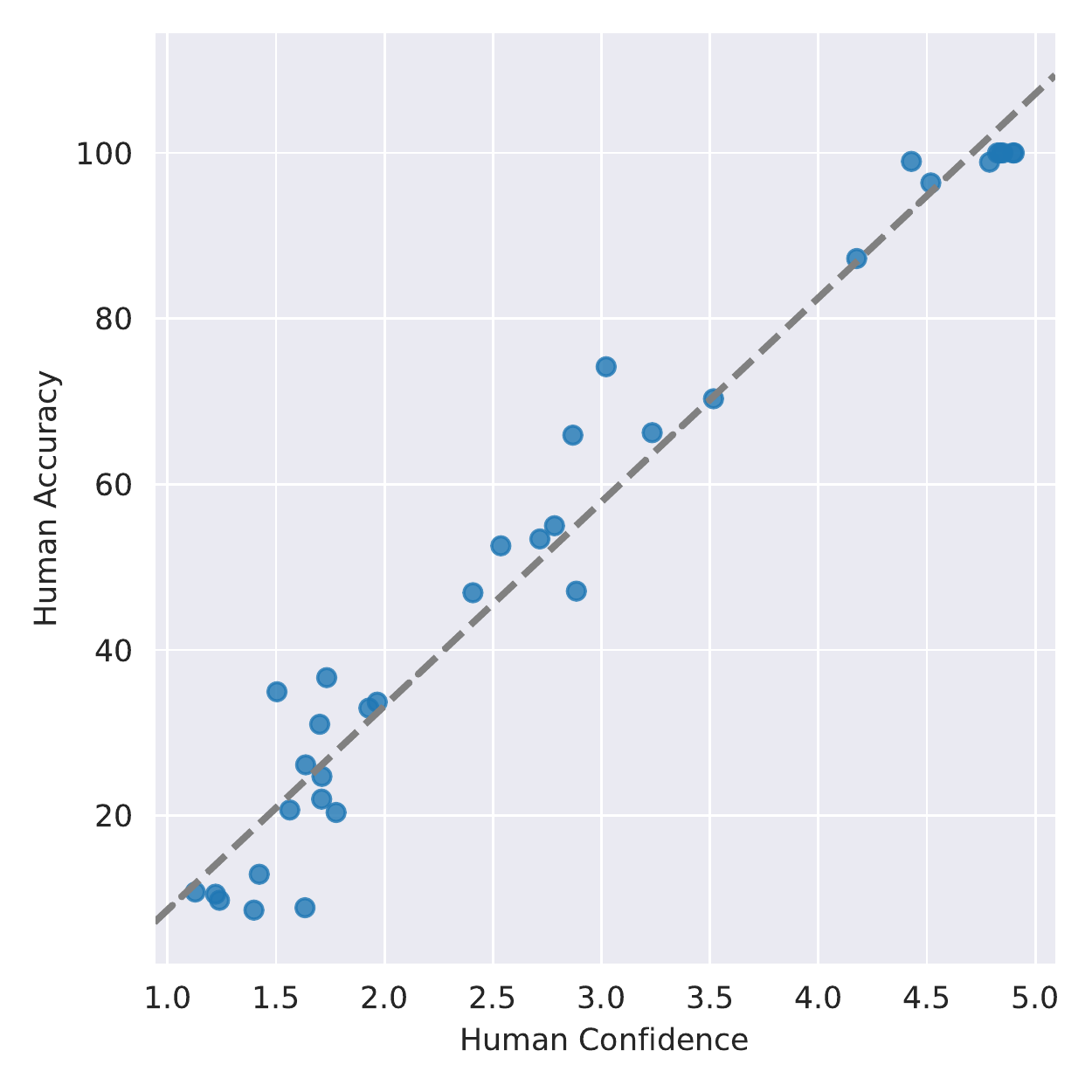}
  \end{center}
  \caption{We found a linear correlation between human confidence scores and human accuracy. Humans are more confident of their performance on the easier transforms where they perform better. Networks show no such trend (Figure \ref{supfig:confidencevsaccnetworks}).}
  \label{fig:confidencevsacchumans}
\end{wrapfigure}
% *** show/talk about the relationship between confidence and accuracy - DONE

We analyzed the difference between the human and machine performance using multiple statistical tests. We tested both absolute performance on the same set of images and the observers' confidence on these images. We used paired \textit{t-}test statistic with 3 degrees of freedom (number of independent variables in our transform) to analyze the difference between networks and humans and found the difference between their performance to be significant (for numbers and transform specific tests, please see Table \ref{table:ttestcorrresponses}). We further used the \textit{Pearson product-moment correlation} to see if the responses were correlated. We found the responses to be only weakly correlated in case of ResNet101, owing to its greater capacity and its performance to be marginally above chance in cases where the other two networks completely give up (for numbers and transform specific tests, please see Table \ref{table:ttestcorrresponses}). We also ran an \textit{Ordinary Least Squares} (OLS) regression between human and network responses, and found similar results (Tables \ref{tab:olsalldata}, \ref{tab:olswithingrid}, \ref{tab:olslocalstructure}, \ref{tab:olssegwithin}, \ref{tab:olssegall}).

To further examine our question about difference in strategy used by humans and machines for solving object recognition task, we statistically analyzed the confidence scores on same images classified by humans and networks. We found the \textit{t-}test statistic to be consistent with our hypothesis about the two being different. (For numbers and transform specific tests, please see Table \ref{table:ttestcorrconfidence}.) The correlation coefficient shows an overall negative correlation between the networks and humans (for numbers and transform specific tests, please see Table \ref{table:ttestcorrconfidence}). While VOneResNet50 shows a non-negative correlation, it is not statistically significant. VOneResNet50 also performed lowest overall. We saw a linear trend in relationship between confidence and accuracy for humans (Figure \ref{fig:confidencevsacchumans}). On tasks where humans performed with a higher accuracy, the confidence scores were high as well. We found the correlation coefficient for this trend to be over 98\%, while the correlation between network confidences and their responses was well below 50\%. 

% below can be a part of limitations if including one.
We plotted saliency maps from 0-shot experiments (please see \textsection\ref{app:saliency}) because including them as part of the training process could introduce additional parameters which could potentially affect our analysis. We also calculated confidence score for networks as described in \cite{gal2016dropout} for a more equitable comparison to the human confidence that we collected in our psychophysics studies. Visualizing the weights of layers of these networks, however, does not show much and would not be a helpful comparison, given that our human experiments do not involve the use of eye-tracking devices or fMRI/EEG techniques. These are left for future studies.

\vspace{4mm}
\label{subsec:transformranking_coarse}
%  *** to be re-written
\noindent\textbf{Ranking of Transforms } 
We saw a linear correlation between human performance and human confidence scores (Figure \ref{fig:confidencevsacchumans}) and found that while they are related for humans, no such trend exists for machines. We next wanted to analyze human performance at a transform-level to compare the relationship across different transforms. While our transforms might look unrelated, they can be recreated by traversing a 3-dimensional space of independent variables, namely i) block size, ii) shuffle probability and iii) moving the block to another location or not. 
We calculated a mean over human and network performance across individual transforms and ranked them in the order of decreasing human performance. Except baselines (no transforms), we did not find both humans and networks agreeing on assigning the same rank to any of the transforms. We found ResNet50 and ResNet101 to agree on most transforms, with the average performance also closely related. VOneResNet50 showed the most uneven trend compared to both humans and ResNets (Figure~\ref{fig:transform_ranking_coarse}), further illustrating its differences on a behavioural level. We present individual rankings per transform for both humans and machines (Table \ref{suptab:transformranking_coarse}). Recall from figures \ref{fig:localstructure} and \ref{fig:gridwithingrid} that variance between peak and average machine performance (not shown explicitly in the figure) as a function of block size is significantly high, while the variance between peak and average human performance (shown as horizontal gray line) is at the center of the range. This behaviour further underscores differences between strategies used by humans and networks to solve our transforms. We present an analysis of parameter-level ranking for transforms in \textsection\ref{supsec:transformranking_fine}.

% \begin{figure}
%   \centering \includegraphics[width=1\textwidth]{figures/ranking.pdf}
%   \caption{Average performance per transform for both humans and networks, sorted by human performance in descending order. Average performance for humans is shown as bars while that for networks is overlaid as lines. None of the networks agree with humans on a transform ranking.}
%   \label{fig:transform_ranking_coarse}
% \end{figure}

% *** motivate about how independent variables can be changed to traverse the axes to create different transforms. Try to create a 2D axes diagram of the things that can be altered to get all the transforms.

\section{Discussion and Conclusion}
% summarize the main results and contributions
% how does it link with human vision
% what might be the causes of discrepancies in the results and what can be concluded from them.
% talk about the claims of VOneNet simulating V1 in their block. Refute gracefully.
% link with the histogram image if possible to talk about how the color intensity distribution might affect performance (weak argument)?

Our work is inspired by the robustness of the human visual system in performing object recognition in the presence of extreme image distortions. We believe that humans use inductive biases and prior knowledge about the world to quickly switch between, or combine, bottom-up and top-down cues from the image features. Primate visual systems have feedback to better understand the visual scenes while most object recognition networks rely on their feedforward behaviour. Recent studies have highlighted the importance of recurrence to compete or exceed network performance on tasks that seem easy for humans \cite{NEURIPS2021linsley, pathtracker}.

Unlike initial layers of CNNs that learn edges, contours, and textures, humans rely on an abstract concept of ``object" representation and further add individual features linking it with an object's category. The abstract concept of ``an object" helps humans to learn about the characteristics of a given object and link it with the accompanying information about its environment. This happens at various levels of the human visual system~\cite{carandini2005we}. During an object's interaction with the environment, humans treat the object (independent of the class) as a whole individual entity, as opposed to parts of it interacting separately~\cite{allison1994human, martin2016grapes, keil2010feature, moon2021integralaction, zmigrod2013feature, levi1997feature}. (The representation of the object being referred to here is atomic in nature. For entities with moving parts, individual parts can be treated as individual objects.) Most work in data transformation is on the augmentation side, wherein the added noise changes the pixel values. We wanted to keep the absolute pixel values intact in our transforms. Our transformations aim to test the understanding of ``objectness" for the popular networks, swinging to the extreme ends of image manipulations.

Our results highlight that while CNNs learn representations in a feature specific manner, largely discounting the characteristic properties of the underlying object, humans try to learn the knowledge of features, building on top of objects~\cite{peronamalikscale, koenderink1984structure, koenderink2021structure}. We find that networks are more affected by our segmentation transforms compared to our block transforms, further indicating their disconnect with human-like behaviour. Networks have learned to solve tasks with noise as part of their training procedures to handle controlled adversarial attacks \cite{learningtoseebylookingatnoise, vonenet}, but struggle when the control is taken away. We filter humans and machines on the adversarial object recognition task, and are not creating systems that can break captchas \cite{noury2020deep, lin2018chinese}. We believe our work could be a step in that direction.  
% *** talk about what the control is.

We show that machines perform better than humans on our ``hard" transforms, while struggle to perform at par with humans on the ``easy" transforms from a human perspective. We also show that this performance is highly correlated with confidence of selecting an object class for humans, while it is random at best for networks, highlighting the difference in strategies used by the two. Recent work on explaining V1 variance and building neural network blocks that simulate the neurophysiological data from visual cortex show promising results on controlled adversarial attacks, but still need more work to behaviourally perform like humans \cite{vonenet, dapello2021neural}. We show that the ranking at which these networks perform the task are very different from humans even at a coarse level. Recent work has applied random noise to pixels and intermediate layers to improve robustness to adversarial attacks \cite{liu2018towards}. Including stochasticity to peripheral models has been proposed as a promising solution to learning more human-like representations \cite{dapello2021neural}. We believe that robustness to such attacks should come from both input transformations and network architecture \cite{geirhos2018generalisation}.  

We demonstrate that human visual system employs more robust strategies in certain instances to solve the object recognition task, and highlight statistical differences in those strategies when compared to machines. Our novel transforms highlight a blind spot for the controlled adversarial training of networks. We hope these transforms can help with development/training of robust architectures simulating tolerance of primate visual system to deal with extreme changes in visual scenes often found in everyday settings.

% can also talk about how networks have taken away appearance from tracking and we envision similar robustness for this field.

\section{Limitation and Future Work}
We used Imagenette \cite{imagenettegit}, a subset of the larger Imagenet dataset \cite{imagenet} with 10 distinct and unrelated classes, due to compute limitations. We think using a larger subset could lead to more stable results, but will not affect the overall differences in patterns observed. We also asked for the confidence score from humans instead of calculating attention maps using fMRI/EEG techniques or using eye-tracking devices, due to limitations with participant recruiting and lack of an appropriate experimental infrastructure. We ran our human experiments in a standard way \cite{NEURIPS2021linsley, frank2017validating} that should not affect the overall trends observed. We additionally filtered the participant responses with catch trials and median absolute deviation \cite{medianabsolutedev}. Not having attention data from humans limits our ability to correlate attention maps or feature weights from networks at a pixel-level.

\backmatter

\bmhead{Supplementary information}

Yes
\bmhead{Acknowledgments}

The authors would like to thank Research Computing at Northeastern University for their storage and computational services. GM is a visiting student at Brown University and would like to thank Center for Computation and Visualization, Brown University and Paulo Baptista for help with computational resources. GM would also like to thank Sobia Shadbar for discussions about the statistical testing. GM is also affiliated to Labrynthe Pvt. Ltd., New Delhi, India. 
\section*{Declarations}

\begin{itemize}
\item Funding: GM was supported by a teaching fellowship from Khoury College at Northeastern University.
\item Conflict of interest/Competing interests: GM is affiliated to Labrynthe Pvt. Ltd., but this work will not directly benefit Labrynthe.
\item Ethics approval: IRB\#: 22-10-09, dated Oct 11, 2022 from Northeastern University
\item Consent to participate: Yes.
\item Consent for publication: Yes.
\item Availability of data and materials: Not available directly, but used for publication in aggregate form.
\item Code availability: No.
\item Authors' contributions: GM conceptualized the study. GM and DC ran network experiments. GM ran human studies and analyzed the data. GM, DC and EM wrote the paper.
\end{itemize}

% \noindent
% If any of the sections are not relevant to your manuscript, please include the heading and write `Not applicable' for that section. 

% %%===================================================%%
% %% For presentation purpose, we have included        %%
% %% \bigskip command. please ignore this.             %%
% %%===================================================%%
% \bigskip
% \begin{flushleft}%
% Editorial Policies for:

% \bigskip\noindent
% Springer journals and proceedings: \url{https://www.springer.com/gp/editorial-policies}

% \bigskip\noindent
% Nature Portfolio journals: \url{https://www.nature.com/nature-research/editorial-policies}

% \bigskip\noindent
% \textit{Scientific Reports}: \url{https://www.nature.com/srep/journal-policies/editorial-policies}

% \bigskip\noindent
% BMC journals: \url{https://www.biomedcentral.com/getpublished/editorial-policies}
% \end{flushleft}

\begin{appendices}

\clearpage
\setcounter{figure}{0}
\setcounter{table}{0}
\setcounter{section}{0}

\makeatletter 
\renewcommand{\thesection}{S\@arabic\c@section}
\renewcommand{\thefigure}{S\@arabic\c@figure}
\renewcommand{\thetable}{S\@arabic\c@table}
\makeatother

\begin{center}

             \Large \textbf{Extreme Image Transformations Affect Humans and Machine Differently\\-- Supplementary Information --}

\end{center}

\section{Human Experiments Setup}
\label{sec:human_setup}
We recruited 32 participants using Cloud Research's Connect platform for our psychophysics study, approved by Northeastern University's IRB (\#22-10-09). The experiment was not time bound and could be completed at participants' own pace. The experiment was designed to take an average of 20-25 minutes. We compensated participants with a pro-rated minimum wage price of \$8 for their time. People with shorter and longer trial times were not compensated lesser or higher.  We recorded the reaction time for all trials. After every trial, participants were redirected to a screen confirming their submission. They could continue by clicking the ``Continue" button or pressing the spacebar. They were automatically redirected from the confirmation screen to the next screen in 2000ms. We also showed a ``rest screen" after completion of every 10 trials, with a progress bar. The rest screen was shown only during main trials and not during practice trials. The time on rest screen was not recorded. 

\paragraph{Experiment design} At the beginning of the experiment, the participants were shown an information screen guiding them about the significance of the experiment and what needs to be done. They could then click ``Continue", which showed an instruction modal pop-up with instructions about what to do. They could view this instruction modal anytime during the experiment by clicking the button ``Instructions" in the top right corner of their screens. 

Participants were shown an image (baseline or transformed) along with ten object classes from the Imagenette dataset. They were asked to identify the object in the image and select the option closest to what they thought the object in the image was. They were also asked to rate their level of confidence on a scale of 1 through 5, ranging from least to most confident. They were given a feedback on their response in the form of correct or incorrect during practice trials but not during the main test trials. Each trial screen also had a short excerpt of instructions. Participants were shown a total of 11 practice trials and 102 test trials.

The class frequency of trial images was not explicitly balanced. In both practice and test trials, we did not focus on the object class as such, but rather on the transforms. We generated multiple images for every transform-hyperparameter pair across multiple classes and randomly selected 3 images for every transform-hyperparameter pair across all classes. After the selection, the same 102 images were shown to both networks and humans. The order of the display of images was randomly shuffled for both.

\paragraph{Software setup} The experiment used Python Flask for backend scripts and logic, and HTML, Bootstrap CSS framework and JavaScript for frontend. The form submission through keys and automatic redirections were done using jQuery on the user side. The server was run on HP Z200 workstation using 1 Intel(R) Xeon(R) CPU and 16 GB RAM.

Imagenette has 3-channel images of $320\times 320$ pixel resolution. The images were sampled from test set at the original resolution for showing to human participants. 

\paragraph{Filtering criteria } Data from cloud sourcing platforms can be noisy, with people identifying shortcuts to either quickly zip through the experiments or take very long on each trial. To eliminate such biases from the data, we excluded participants who either took a very long time, or failed on the baseline catch trials. To achieve our earlier filtering criteria, we removed participants for whom the response time did not exceed 2 median absolute deviations below the median, $median(X) - 2 * MAD(X)$, where MAD = median
absolute deviation \cite{medianabsolutedev}, a robust alternative to mean and standard deviation for identifying outliers. 

Out of the 32 participants we recruited, 2 participants were filtered out by the above criteria. We removed all trials for the filtered out participants. Additionally, we recorded the accuracy of participants in practice trials and found all of them to perform with over 70\% accuracy. We had also included catch trials in form of baseline and color flatten images, which are the easiest and hardest to classify respectively. None of the participants were found to be inattentive on the catch trials.

\section{Statistical Analysis of Human and Network Data}
\label{supsec:statanalysis}
% *** show table of difficulty ranking in supplementary as well
To test the significance of human responses compared to the data returned from the networks, we ran multiple statistical tests, trying to understand if the performance was significant, and if the strategies used by the networks were similar in any way. We used \textit{t-}test statistic, correlation coefficient and Ordinary Least Squares regression (OLS) to fit human and network data. We used \texttt{statsmodels}~\cite{seabold2010statsmodels} library in python for our analysis. For \textit{t-}test, we considered 3 degrees of freedom based on the number of independent variables that could be altered to get our transforms -- block size, probability of shuffle, and moving the block to another position or not.

\section{Ranking of Transforms}
\label{supsec:transformranking_fine}
Our transforms are based on three independent variables -- i) Block size (or number of segments in case of segmentation shuffles), ii) Probability of individual pixel shuffle and iii) Moving the block/region to another location or not. Traversing this 3-dimensional space leads to a wide variety of variations in the visual perception of objects for humans and machines. To link it all together, we ranked the transforms by sorting collective human performances on the transformation-parameter pairs. We calculate an overall ranking of all the transforms in Table \ref{tab:humanresponseblocktransforms} and plot it based on their probability and block sizes (10\% accuracy is treated as chance). Difficulty rank is calculated by $100 - accuracy$ across all images in the respective transformation-parameter pair (higher rank is harder). For a discussion about coarse transform-level ranking, please see \textsection\ref{subsec:transformranking_coarse}.

\begin{figure}
  \centering \includegraphics[width=1\textwidth]{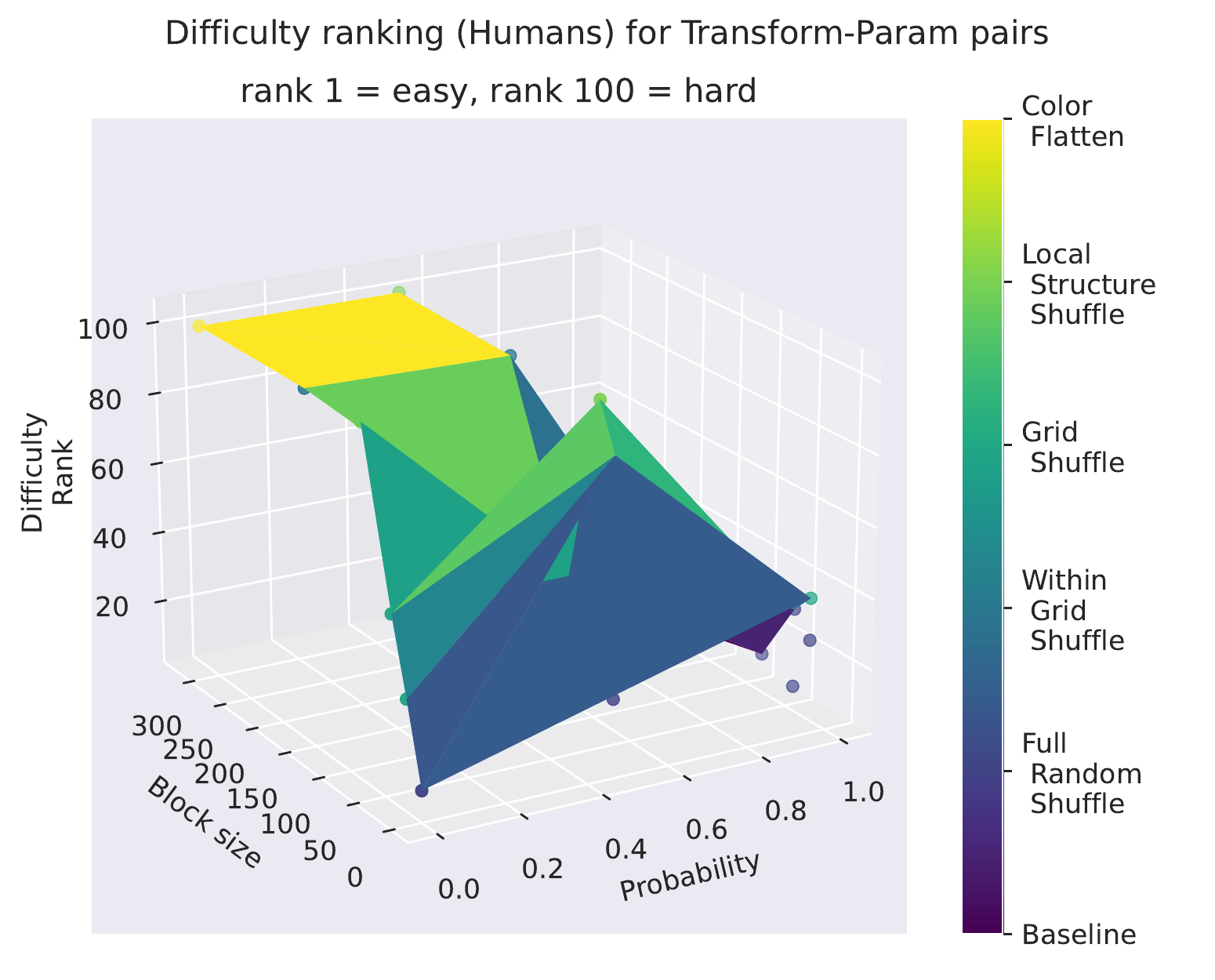}
  \caption{Ranking of Transforms: Difficulty rank as a function of block size and probability for human observers. Higher rank is harder for humans. The surface looks like a bird flapping wings, indicating lower difficulty for a few cases, but shows an overall increase in difficulty with a change in independent variables. See text for details}
  \label{fig:rankingsurface}
\end{figure}

The influence of independent variables on the performance of humans is shown by the surface in Fig \ref{fig:rankingsurface}. The shape of surface looks like that of a bird flapping its wings. We found the performance to be inversely related to the increase in probability of shuffle, and directly related to an increase in block size, in line with what has been observed in the field \cite{nguyen2015deep, xie2020self, liu2022towards, chen2021robust, shen2020noise, kaneko2020noise} We observed an opposite trend in case of networks. The conditions where humans performed well, were hard for networks and vice-versa (Table \ref{table:ttestcorrconfidence}). Human responses are treated as independent variables. 

While we observe a global upward trend in difficulty of transforms, we also observe certain outliers where complexity decreases for certain transforms. This is specially true in cases like Within Grid Shuffle (probability = 0.5), Local Structure Shuffle (probability = 0.5 and block size = 160) [shown in turquoise and green], where the larger block sizes aid in feature identification. 

\begin{figure}
  \centering \includegraphics[width=.3\textwidth]{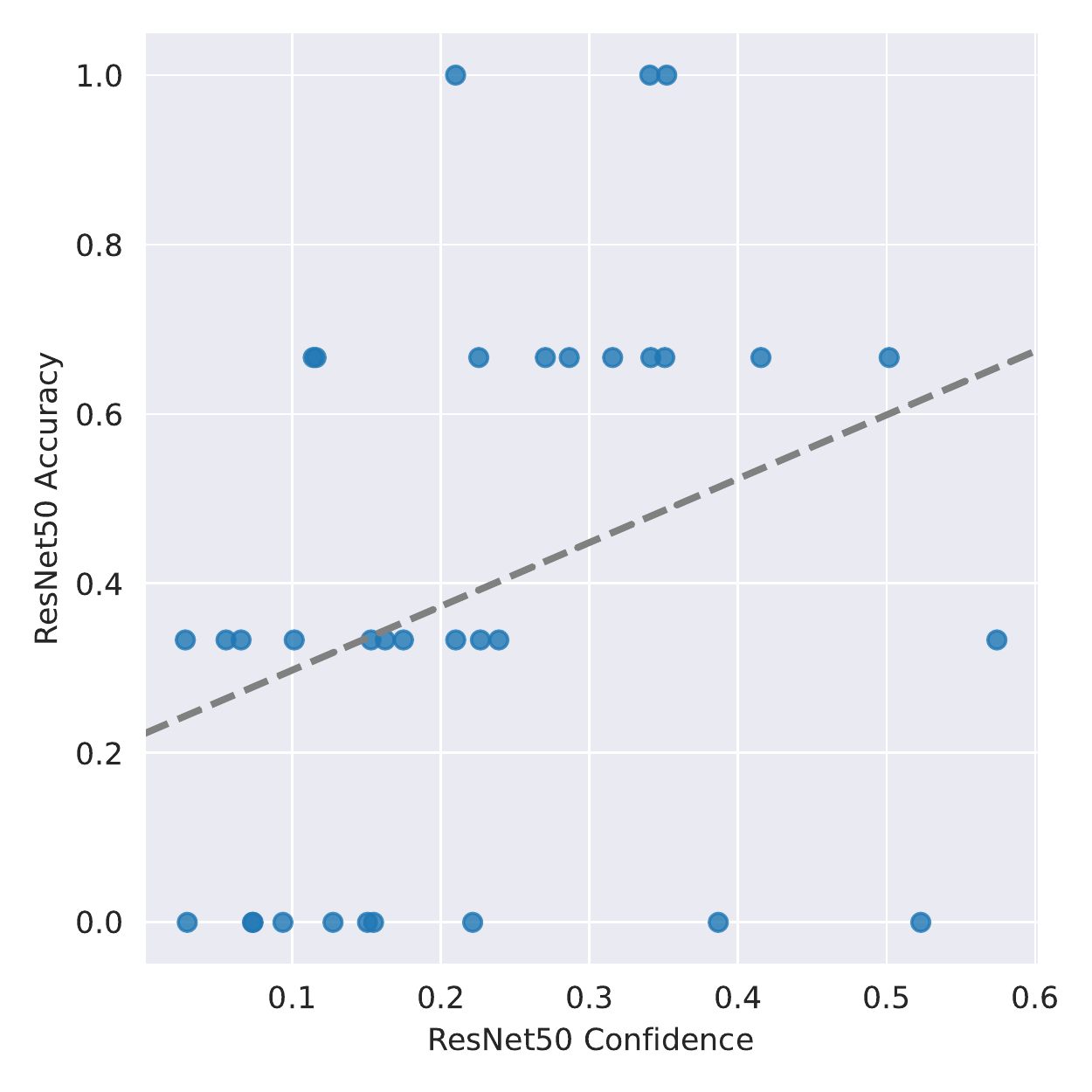}
  \centering \includegraphics[width=.3\textwidth]{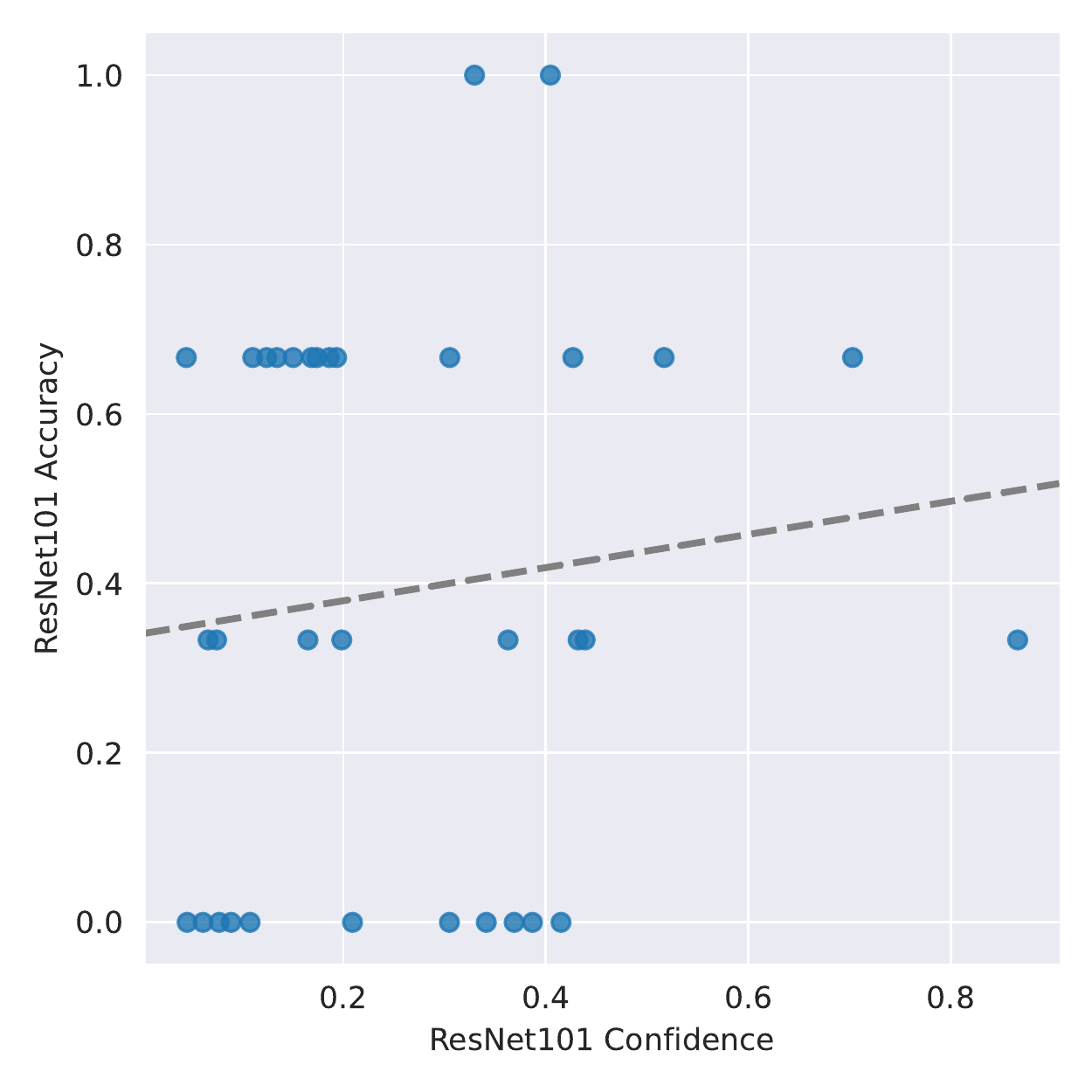}
  \centering \includegraphics[width=.3\textwidth]{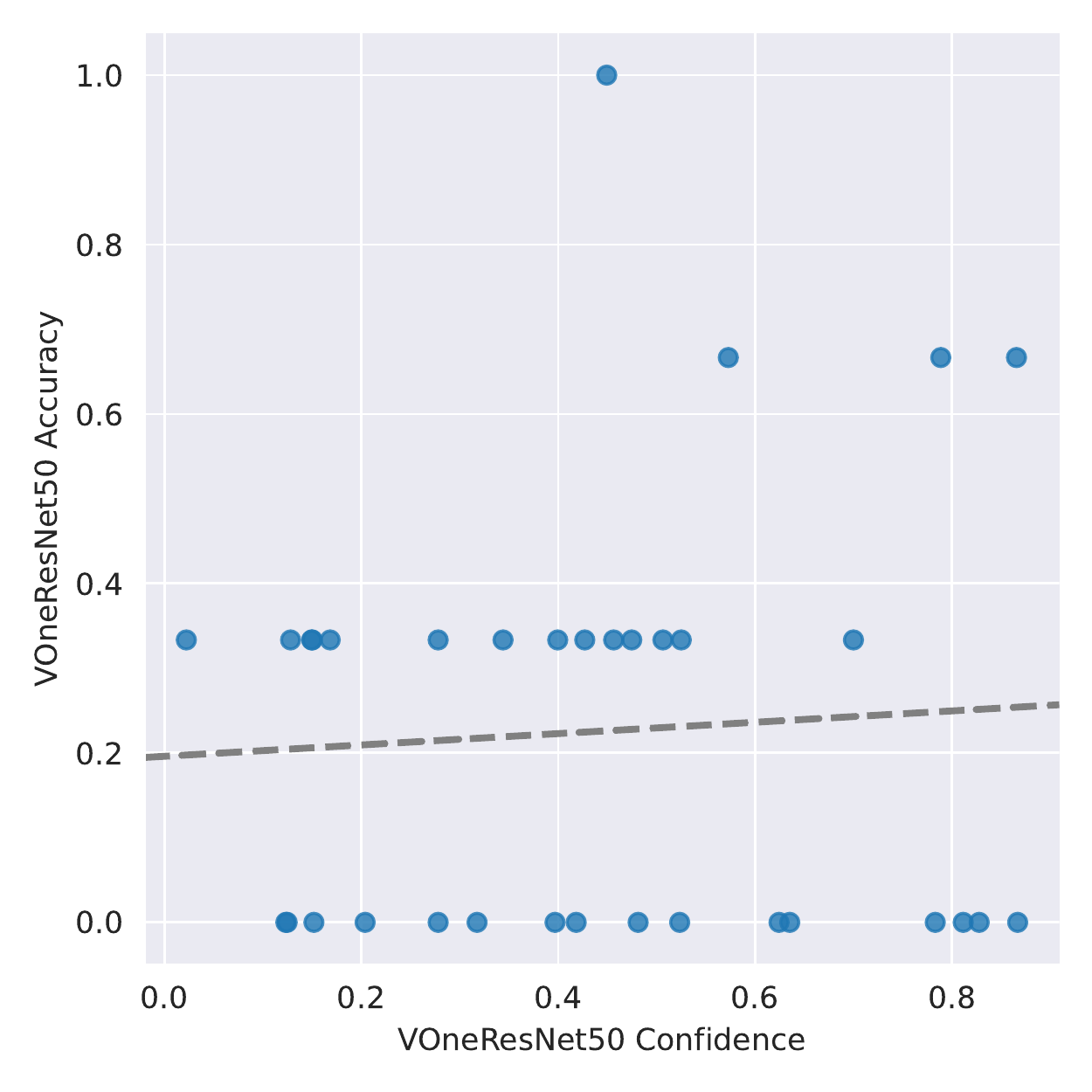}
  \caption{Correlation between network confidence and accuracy. (L-R) ResNet50, ResNet101, VOneResNet50. We see that the trend is not linear (as in case of humans, Fig \ref{fig:confidencevsacchumans}).}
  \label{supfig:confidencevsaccnetworks}
\end{figure}

\paragraph{Correlation in network responses}
Similar to the correlation analysis between human accuracy and human confidence (Figure \ref{fig:confidencevsacchumans}), we tried linear curve fitting for network accuracy v/s network confidence. While humans show a clearly linear trend, networks do not show any such correlations (Figure \ref{supfig:confidencevsaccnetworks}).

\begin{figure}
  \centering \includegraphics[width=0.8\textwidth]{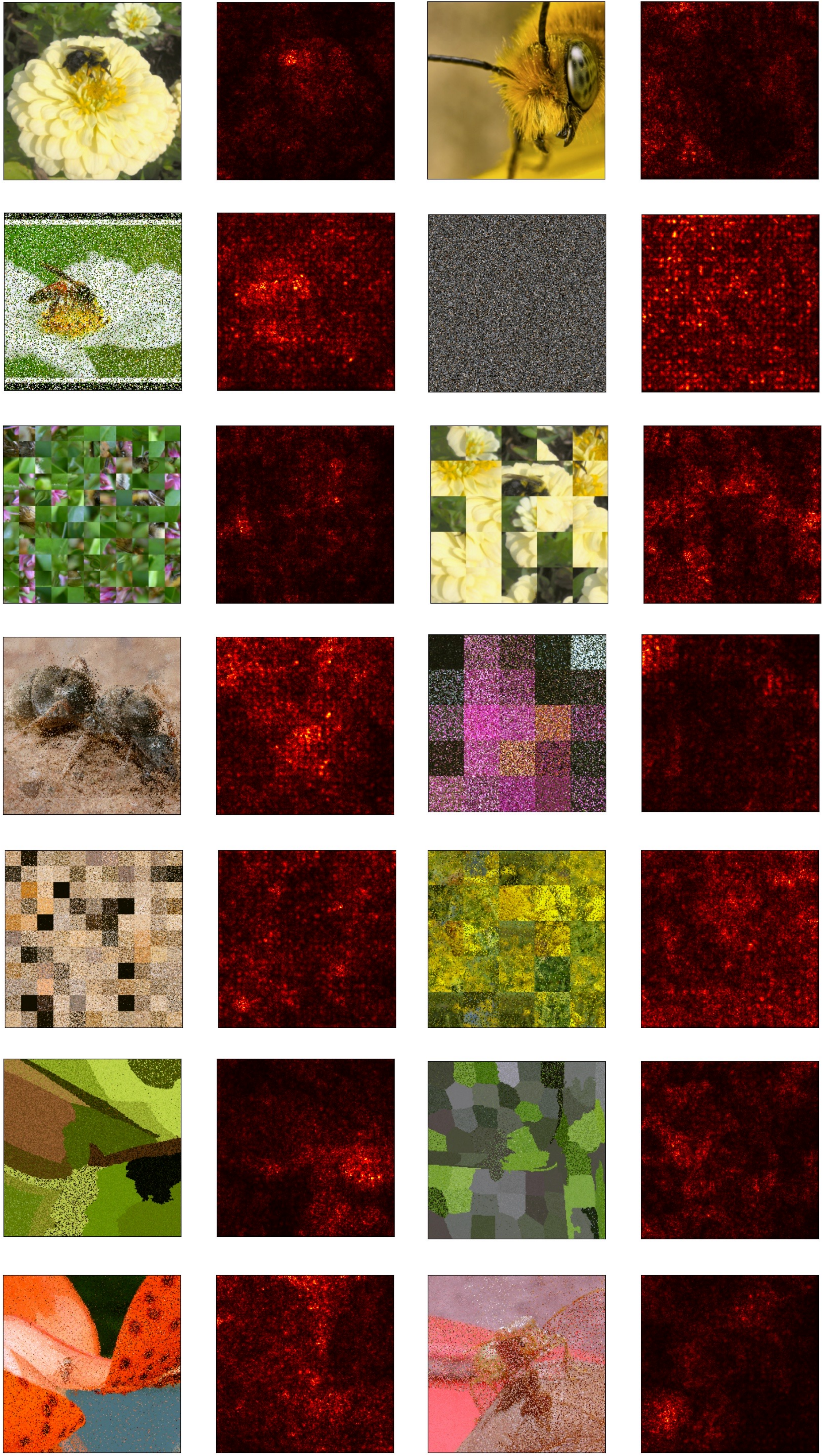}
  \caption{Extreme Image Transformed images and saliency maps from Imagenet pretrained ResNet50 evaluated on Hymenoptera dataset. Images \textbf{L-R and T-B} are baseline, baseline, Full Random Shuffle($p=0.5$ and $p=1.0$), Grid Shuffle($b=20$ and $b=40$), Within Grid Shuffle($b=20,p=0.5$ and $b=40,p=1.0$), Local Structure Shuffle($b=20,p=1.0$ and $b=40,p=0.5$), Segmentation Displacement Shuffle($b=16$ and $b=64$), and Segment Within Shuffle($b=8,p=0.5$ and $b=16,p=0.5$). The saliency maps clearly show that ResNet50 focuses mostly on the contrast of patches and does not explicitly attend to the object in focus.}
  \label{fig:saliencyresnet50}
\end{figure}

\begin{figure}
  \centering \includegraphics[width=0.8\textwidth]{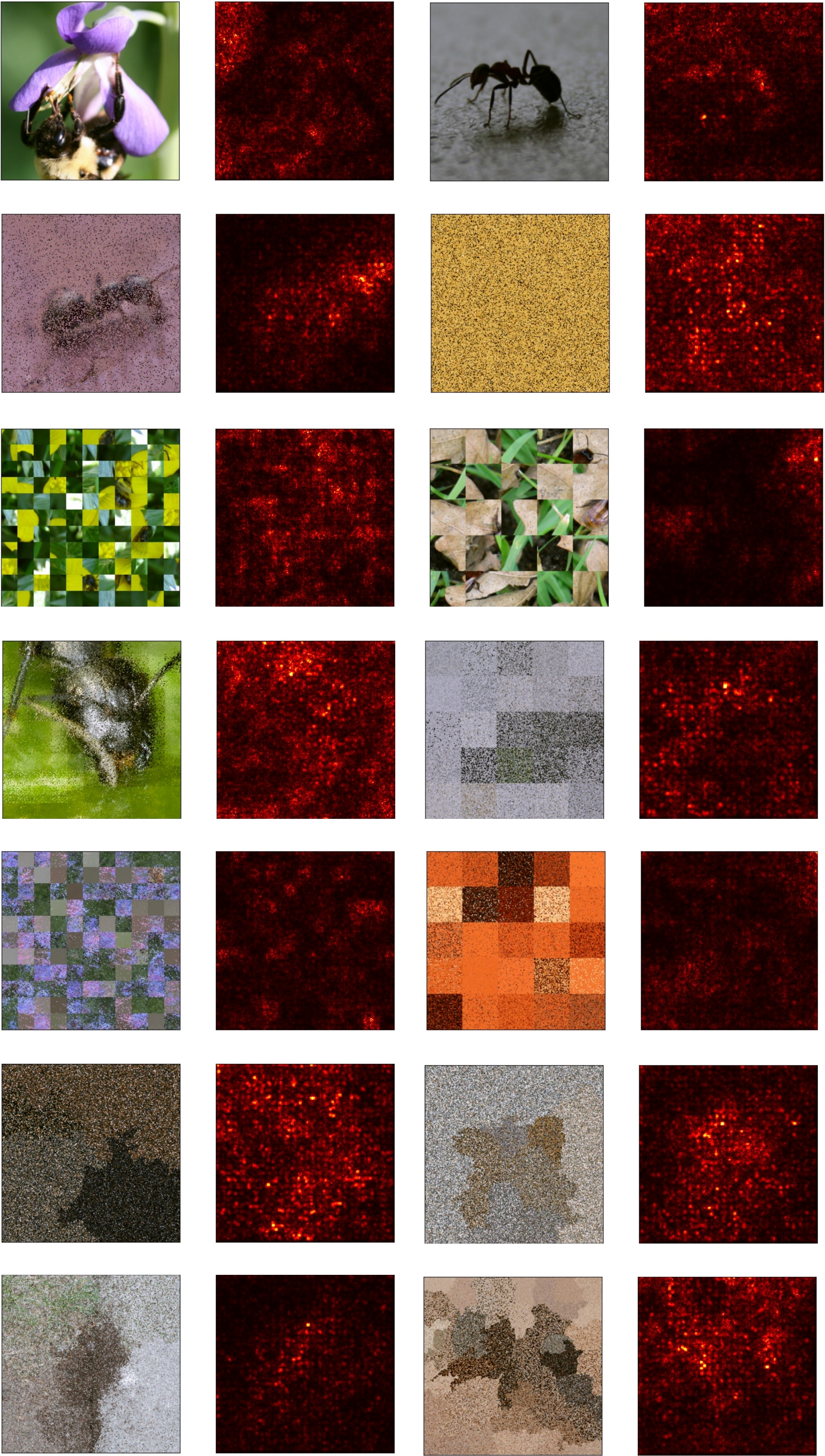}
  \caption{Extreme Image Transformed images and saliency maps from Imagenet pretrained ResNet101 evaluated on Hymenoptera dataset. Images \textbf{L-R and T-B} are baseline, baseline, Full Random Shuffle($p=0.5$ and $p=1.0$), Grid Shuffle($b=20$ and $b=40$), Within Grid Shuffle($b=20,p=0.5$ and $b=40,p=1.0$), Local Structure Shuffle($b=20,p=0.5$ and $b=40,p=1.0$), Segmentation Displacement Shuffle($b=8$ and $b=64$), and Segment Within Shuffle($b=8,p=0.5$ and $b=64,p=1.0$). The saliency maps clearly show that ResNet101 focuses mostly on the contrast of patches and does not explicitly attend to the object in focus. The area covered in saliency maps is higher than that compared to ResNet50, mostly due to the higher number of parameters in the network.}
  \label{fig:saliencyresnet101}
\end{figure}

\paragraph{Insights into how networks classify images}
\label{app:saliency}
To further substantiate our claim that networks cannot generalize to novel objects without the need for finetuning --- while humans can recognize and generalize in the wild --- we performed 0-shot experiments with ResNet50, ResNet101 and VOneResNet50. We started with Imagenet~\cite{imagenet} weights for all 3 networks and evaluated them on the Hymenoptera dataset~\cite{munoz2010hymenoptera, elsik2016hymenoptera} containing bees and ants. We found that none of the 3 networks were able to classify the bees and ants correctly on baseline and transforms alike. We instead found that networks worked well with the low-level feature of color contrasts and showed high activity where the objects stood out from the background, as shown by their respective saliency maps (Figures~\ref{fig:saliencyresnet50}, \ref{fig:saliencyresnet101}). In transforms where the contrasts between foreground and background are not strong, or where the pixels are more randomly shuffled, the activity in these networks is more evenly distributed. In case of segmentation shuffle transforms, the networks do not seem to trace the contours of segmentation.

\begin{table}[]
\caption{Transform-level Ranking for Humans and Networks, sorted by Human performance. Higher rank is harder.}
    \label{suptab:transformranking_coarse}
    \centering
\begin{tabular}{lcccc}
\toprule
\textbf{Transform} & \textbf{Human} & \textbf{ResNet50} & \textbf{ResNet101} & \textbf{VOne} \\   
\midrule
                            %   p       size        50          101         vone
\rowcolor[gray]{.9} Baseline                     & 1    & 1     & 1     & 1        \\
Full Random Shuffle                              & 2    & 2     & 2     & 3         \\ 
\rowcolor[gray]{.9} Grid Shuffle                 & 3    & 6     & 6     & 8         \\
Within Grid Shuffle                              & 4    & 7     & 7     & 6       \\
\rowcolor[gray]{.9} Local Structure Shuffle      & 5    & 4     & 3     & 2       \\ 
Segmentation Displacement Shuffle                & 6    & 5     & 5     & 5        \\ 
\rowcolor[gray]{.9} Segmentation Within Shuffle  & 7    & 8     & 7     & 7         \\
Color flatten                                    & 8    & 3     & 4     & 4         \\
\bottomrule
\end{tabular}
\end{table}

\begin{table}[ht]
\caption{Test accuracy for humans and networks trained on Imagenette dataset with Block transforms}
\label{tab:humanresponseblocktransforms}
\begin{tabular}{lcccccc}
\toprule

\multirow{2}{*}{\textbf{Transform}}  & \multirow{2}{*}{\textbf{P}} & \multirow{2}{2em}{\textbf{Grid Size}} & \multicolumn{4}{c}{\textbf{Accuracy (in \%)}}\\  \cmidrule{4-7}
& & &\textbf{ResNet50} & \textbf{ResNet101} & \textbf{VOne} & \textbf{Human}   \\  
\midrule

                            %   p       size        50          101         vone
\text{Baseline}             &           &           & 66.67     & 33.33     & 0         & 100      \\ \cmidrule{1-7}
\text{Full Random Shuffle}       & 0.5       &           & 66.67     & 66.67     & 0         & 100    \\ 
                            & 0.8       &           & 0         & 0         & 66.67     & 64.06          \\ 
                            & 1.0       &           & 33.33     & 0         & 33.3      & 12.82    \\ \cmidrule{1-7}
\text{Grid Shuffle}         &           & 20x20     & 0         & 0         & 0         & 32.99    \\
                            &           & 40x40     & 33.33     & 33.33     & 33.33     & 53.42     \\ 
                            &           & 80x80     & 66.67     & 66.67     & 33.33     & 100     \\ 
                            &           & 160x160   & 66.67     & 66.67     & 66.67     & 100     \\ \cmidrule{1-7}
\text{Within Grid Shuffle}  & 0.5       & 20x20     & 33.33     & 33.33     & 0         & 87.25     \\
                            &           & 40x40     & 66.67     & 66.67     & 0         & 100    \\ 
                            &           & 80x80     & 0         & 0         & 0         & 47.11     \\ 
                            &           & 160x160   & 66.67     & 66.67     & 66.67     & 98.89     \\ \cmidrule{2-7} 
                            & 1.0       & 20x20     & 33.33     & 33.33     & 33.33     & 24.74     \\
                            &           & 40x40     & 66.67     & 66.67     & 66.67     & 31.03     \\ 
                            &           & 80x80     & 33.33     & 66.67     & 33.33     & 34.96    \\ 
                            &           & 160x160   & 100       & 66.67     & 0         & 9.78    \\ \cmidrule{1-7}
\text{Local Structure Shuffle}   & 0.5       & 20x20     & 1         & 1         & 33.33     & 20.39    \\
                            &           & 40x40     & 66.67     & 66.67     & 1         & 55    \\ 
                            &           & 80x80     & 0         & 33.33     & 0         & 46.91    \\ 
                            &           & 160x160   & 66.67     & 66.67     & 33.33     & 70.33     \\ \cmidrule{2-7} 
                            & 1.0       & 20x20     & 1         & 1         & 0         & 36.67      \\
                            &           & 40x40     & 0         & 0         & 0     & 8.89    \\ 
                            &           & 80x80     & 33.33     & 33.33     & 0         & 12.93    \\ 
                            &           & 160x160   & 33.33     & 0         & 33.33     & 8.6      \\ \cmidrule{1-7}
\text{Color Flatten}        &           &           & 66.67     & 66.67     & 33.33     & 10.78    \\ 
\bottomrule

\end{tabular}
\end{table}

\begin{table}[ht]
\caption{Test accuracy for humans and networks trained on Imagenette dataset with Segmentation transforms}
\label{tab:humanresponsesegmentationtransforms}
\begin{tabular}{p{10mm}cp{13mm}cccc}
\toprule
% todo fix segments, formatting looks weird
\multirow{2}{*}{\textbf{Transform}}  & \multirow{2}{*}{\textbf{P}} & \multirow{2}{2em}{\textbf{Segments}} & \multicolumn{4}{c}{\textbf{Accuracy (in \%)}}\\  \cmidrule{4-7}
& & &\textbf{ResNet50} & \textbf{ResNet101} & \textbf{VOne} & \textbf{Human} \\  
\midrule

                            %   p       size        50          101         vone
\multicolumn{1}{c}{\multirow{2}{8em}{Segmentation Displacement Shuffle}}  &      & 8        & 33.33     & 33.33     & 33.33 & 22     \\
                             &           & 16       & 0         & 0         & 0     & 20.69   \\ 
                             &           & 64       & 0         & 0         & 0     & 26.14   \\ \cmidrule{1-7}
\multicolumn{1}{c}{\multirow{2}{8em}{Segmentation Within Shuffle}}              & 0.5       & 8        & 33.33     & 0         & 33.33 & 65.93   \\
                             &           & 16       & 33.33     & 66.67     & 33.33 & 96.39  \\ 
                             &           & 64       & 33.33     & 66.67     & 33.33 & 98.98   \\ \cmidrule{2-7} 
                             & 1.0       & 8        & 0         & 33.33     & 0     & 52.58   \\
                             &           & 16       & 0         & 0         & 0     & 74.19   \\ 
                             &           & 64       & 0         & 0         & 0     & 33.71  \\ 
\bottomrule
\end{tabular}
\end{table}

\begin{table}[ht]
\caption{\textit{t-}test statistic and Correlation Coefficient between human and network responses. Networks are ResNet\textbf{50}, ResNet\textbf{101}, and \textbf{VOne}ResNet50 in that order. Human responses are averaged across all participants for the mentioned transforms. p-values are indicated in the parentheses below the statistic.}
\label{table:ttestcorrresponses}
\begin{tabular}{lcccccc}
\toprule
\multirow{2}{*}{\textbf{Transform}}  & \multicolumn{3}{c}{\bfseries \textit{t-}test Statistic} & \multicolumn{3}{c}{\bfseries Correlation Coefficient} \\ \cmidrule(l){2-4} \cmidrule(l){5-7} 
&\textbf{50} & \textbf{101} & \textbf{VOne} &\textbf{50} & \textbf{101} & \textbf{VOne} \\   
\midrule
                            %   p       size        50          101         vone
\rowcolor[gray]{.9} All transforms               &  -1.57    & -1.46    & -4.55     & 0.17      &0.32   & 0.16     \\
\rowcolor[gray]{.9}                              &  (0.1217) & (0.1217)	&	(0.0002)  & & &\\
Full Random Shuffle                              &  -0.79    & -1.07    & -1.29     & 0.37      & 0.79  & -0.79   \\ 
                                                 &(0.4777) &(0.3459) &(0.2951)  & & &\\
\rowcolor[gray]{.9} Grid Shuffle                 & -1.29     & -1.29    & -1.76     & 0.98      & 0.98  & 0.81   \\
\rowcolor[gray]{.9}                              &(0.2456) &(0.2456) &(0.1307)  & & &\\
Within Grid Shuffle                              & -0.25     & -0.27    & -1.78     & -0.06     & 0.06  & -0.03   \\
                                                 &(0.8044) &(0.7896) &(0.0976)  & & &\\
\rowcolor[gray]{.9} Local Structure Shuffle      & 1.07      & 1.07     & -1.52     & 0.28      & 0.47  & 0.28 \\ 
\rowcolor[gray]{.9}                              &(0.3053) &(0.3053) &(0.6213)  & & &\\
Segmentation Displacement Shuffle                & -1.05     & -1.05    & -1.05     & -0.29     & -0.29 & -0.29   \\ 
                                                 &(0.3987) &(0.3987) &(0.3987)  & & &\\
\rowcolor[gray]{.9} Segmentation Within Shuffle  & -4.22     & -2.32    & -4.22     & 0.73      & 0.77  & 0.73   \\
\rowcolor[gray]{.9}                              &(0.0022) &(0.0319) &(0.0022)  & & &\\
\bottomrule
\end{tabular}
\end{table}

\begin{table}[ht]
\caption{\textit{t-}test statistic and Correlation Coefficient between human and network confidence. Networks are ResNet\textbf{50}, ResNet\textbf{101}, and \textbf{VOne}ResNet50 in that order. Human confidence is averaged across all participants for the mentioned transforms. p-values are indicated in the parentheses below the statistic.}
\label{table:ttestcorrconfidence}
\begin{tabular}{lcccccc}
\toprule
\multirow{2}{*}{\textbf{Transform}}  & \multicolumn{3}{c}{\bfseries \textit{t-}test Statistic} & \multicolumn{3}{c}{\bfseries Correlation Coefficient} \\ \cmidrule(l){2-4} \cmidrule(l){5-7} 
&\textbf{50} & \textbf{101} & \textbf{VOne} &\textbf{50} & \textbf{101} & \textbf{VOne} \\   
\midrule
                            %   p       size        50          101         vone
\rowcolor[gray]{.9}  All transforms                   &  -11.00   & -10.78   & -9.95     & -0.06     &-0.17  & 0.22     \\
\rowcolor[gray]{.9}                              &  (1.0492E-12) & (1.4316E-12)	&	(8.8715E-12)  & & &\\
Full Random Shuffle                              &  -2.77    & -2.68    & -2.67     & 0.33      & -0.74 & 0.65   \\
                                                 &(0.1084) &(0.1152) &(0.1166)  & & &\\
\rowcolor[gray]{.9} Grid Shuffle            & -4.48     & -4.57    & -4.14     & 0.95      & 0.17  & 0.25   \\
\rowcolor[gray]{.9}                              &(0.0202) &(0.0196) &(0.0253)  & & &\\
Within Grid Shuffle                         & -4.77     & -4.58    & -3.90     & -0.11     & -0.02 & 0.43   \\
                                                &(0.0019) &(0.0023) &(0.0057) & & &\\
\rowcolor[gray]{.9} Local Structure  & -6.53     & -6.10    & -5.90     & -0.21     & -0.40 & 0.24 \\ 
\rowcolor[gray]{.9}  Shuffle                            &(0.0002) &(0.0003) &(0.0005)  & & &\\
Segmentation             & -25.56    & -12.14   & -14.23    & 0.68      & -0.04 & 0.01   \\ 
 Displacement Shuffle                                                &(0.0000) &(0.0024) &(0.0010)  & & &\\
\rowcolor[gray]{.9} Segmentation Within     & -7.11     & -7.44    & -7.26     & -0.69     & 0.54  & -0.02   \\
\rowcolor[gray]{.9}         Shuffle                     &(0.0007) &(0.0007) &(0.0000)  & & &\\
\bottomrule
\end{tabular}
\end{table}

% running OLS test for  all
\begin{center}
\begin{table}
\caption{OLS for all data (baseline and transforms)}
\label{tab:olsalldata}
\begin{tabular}{lclc}
\toprule
\textbf{Dep. Variable:}    &        y         & \textbf{  R-squared (uncentered):}      &     0.903   \\
\textbf{Model:}            &       OLS        & \textbf{  Adj. R-squared (uncentered):} &     0.890   \\
\textbf{Method:}           &  Least Squares   & \textbf{  F-statistic:       }          &     69.87   \\
\textbf{Date:}             & Sat, 26 Nov 2022 & \textbf{  Prob (F-statistic):}          &  9.12e-15   \\
\textbf{Time:}             &     13:13:29     & \textbf{  Log-Likelihood:    }          &   -8.5703   \\
\textbf{No. Observations:} &          34      & \textbf{  AIC:               }          &     25.14   \\
\textbf{Df Residuals:}     &          30      & \textbf{  BIC:               }          &     31.25   \\
\textbf{Df Model:}         &           4      & \textbf{                     }          &             \\
\bottomrule
\end{tabular}
\begin{tabular}{lcccccc}
            & \textbf{coef} & \textbf{std err} & \textbf{t} & \textbf{P$> |$t$|$} & \textbf{[0.025} & \textbf{0.975]}  \\
\midrule
\textbf{Humans} &       0.1658  &        0.035     &     4.713  &         0.000        &        0.094    &        0.238     \\
\textbf{ResNet50} &       0.8358  &        0.396     &     2.111  &         0.043        &        0.027    &        1.644     \\
\textbf{ResNet101} &       0.5925  &        0.334     &     1.775  &         0.086        &       -0.089    &        1.274     \\
\textbf{VOneResNet50} &       0.2324  &        0.268     &     0.867  &         0.393        &       -0.315    &        0.780     \\
\bottomrule
\end{tabular}
\begin{tabular}{lclc}
\textbf{Omnibus:}       &  2.006 & \textbf{  Durbin-Watson:     } &    0.975  \\
\textbf{Prob(Omnibus):} &  0.367 & \textbf{  Jarque-Bera (JB):  } &    1.347  \\
\textbf{Skew:}          & -0.229 & \textbf{  Prob(JB):          } &    0.510  \\
\textbf{Kurtosis:}      &  2.139 & \textbf{  Cond. No.          } &     23.6  \\
\bottomrule
\end{tabular}
\end{table}
%\caption{OLS Regression Results}
\end{center}

% running OLS test for  Within Grid Shuffle
\begin{center}
\begin{table}
\caption{OLS for Within Grid Shuffle}
\label{tab:olswithingrid}
\begin{tabular}{lclc}
\toprule
\textbf{Dep. Variable:}    &        y         & \textbf{  R-squared (uncentered):}      &     0.975   \\
\textbf{Model:}            &       OLS        & \textbf{  Adj. R-squared (uncentered):} &     0.950   \\
\textbf{Method:}           &  Least Squares   & \textbf{  F-statistic:       }          &     39.38   \\
\textbf{Date:}             & Sat, 26 Nov 2022 & \textbf{  Prob (F-statistic):}          &  0.00181    \\
\textbf{Time:}             &     13:13:29     & \textbf{  Log-Likelihood:    }          &    3.4419   \\
\textbf{No. Observations:} &           8      & \textbf{  AIC:               }          &     1.116   \\
\textbf{Df Residuals:}     &           4      & \textbf{  BIC:               }          &     1.434   \\
\textbf{Df Model:}         &           4      & \textbf{                     }          &             \\
\bottomrule
\end{tabular}
\begin{tabular}{lcccccc}
            & \textbf{coef} & \textbf{std err} & \textbf{t} & \textbf{P$> |$t$|$} & \textbf{[0.025} & \textbf{0.975]}  \\
\midrule
\textbf{Humans} &      -0.0244  &        0.067     &    -0.366  &         0.733        &       -0.209    &        0.160     \\
\textbf{ResNet50} &      -0.2386  &        0.693     &    -0.344  &         0.748        &       -2.164    &        1.686     \\
\textbf{ResNet101} &      -0.1106  &        0.473     &    -0.233  &         0.827        &       -1.425    &        1.204     \\
\textbf{VOneResNet50} &       1.5680  &        0.568     &     2.760  &         0.051        &       -0.009    &        3.145     \\
\bottomrule
\end{tabular}
\begin{tabular}{lclc}
\textbf{Omnibus:}       &  1.314 & \textbf{  Durbin-Watson:     } &    1.668  \\
\textbf{Prob(Omnibus):} &  0.518 & \textbf{  Jarque-Bera (JB):  } &    0.718  \\
\textbf{Skew:}          &  0.310 & \textbf{  Prob(JB):          } &    0.698  \\
\textbf{Kurtosis:}      &  1.669 & \textbf{  Cond. No.          } &     37.7  \\
\bottomrule
\end{tabular}
\end{table}
%\caption{OLS Regression Results}
\end{center}

% running OLS test for  Local Structure Shuffle
\begin{center}
\begin{table}
\caption{OLS for Local Structure Shuffle}
\label{tab:olslocalstructure}
\begin{tabular}{lclc}
\toprule
\textbf{Dep. Variable:}    &        y         & \textbf{  R-squared (uncentered):}      &     0.992   \\
\textbf{Model:}            &       OLS        & \textbf{  Adj. R-squared (uncentered):} &     0.984   \\
\textbf{Method:}           &  Least Squares   & \textbf{  F-statistic:       }          &     125.4   \\
\textbf{Date:}             & Sat, 26 Nov 2022 & \textbf{  Prob (F-statistic):}          &  0.000187   \\
\textbf{Time:}             &     13:13:29     & \textbf{  Log-Likelihood:    }          &    8.0076   \\
\textbf{No. Observations:} &           8      & \textbf{  AIC:               }          &    -8.015   \\
\textbf{Df Residuals:}     &           4      & \textbf{  BIC:               }          &    -7.698   \\
\textbf{Df Model:}         &           4      & \textbf{                     }          &             \\
\bottomrule
\end{tabular}
\begin{tabular}{lcccccc}
            & \textbf{coef} & \textbf{std err} & \textbf{t} & \textbf{P$> |$t$|$} & \textbf{[0.025} & \textbf{0.975]}  \\
\midrule
\textbf{Humans} &       0.0617  &        0.064     &     0.958  &         0.392        &       -0.117    &        0.240     \\
\textbf{ResNet50} &       0.3270  &        0.695     &     0.471  &         0.662        &       -1.602    &        2.256     \\
\textbf{ResNet101} &       0.5928  &        0.460     &     1.288  &         0.267        &       -0.685    &        1.870     \\
\textbf{VOneResNet50} &       1.0817  &        0.251     &     4.313  &         0.013        &        0.385    &        1.778     \\
\bottomrule
\end{tabular}
\begin{tabular}{lclc}
\textbf{Omnibus:}       &  0.077 & \textbf{  Durbin-Watson:     } &    1.757  \\
\textbf{Prob(Omnibus):} &  0.962 & \textbf{  Jarque-Bera (JB):  } &    0.153  \\
\textbf{Skew:}          &  0.109 & \textbf{  Prob(JB):          } &    0.926  \\
\textbf{Kurtosis:}      &  2.358 & \textbf{  Cond. No.          } &     42.4  \\
\bottomrule
\end{tabular}
\end{table}
%\caption{OLS Regression Results}
\end{center}

% running OLS test for  Seg Within Shuffle
\begin{center}
\begin{table}
\caption{OLS for Segmentation Within Shuffle}
\label{tab:olssegwithin}
\begin{tabular}{lclc}
\toprule
\textbf{Dep. Variable:}    &        y         & \textbf{  R-squared (uncentered):}      &     0.987   \\
\textbf{Model:}            &       OLS        & \textbf{  Adj. R-squared (uncentered):} &     0.961   \\
\textbf{Method:}           &  Least Squares   & \textbf{  F-statistic:       }          &     38.11   \\
\textbf{Date:}             & Sat, 26 Nov 2022 & \textbf{  Prob (F-statistic):}          &   0.0257    \\
\textbf{Time:}             &     13:13:29     & \textbf{  Log-Likelihood:    }          &    4.5261   \\
\textbf{No. Observations:} &           6      & \textbf{  AIC:               }          &    -1.052   \\
\textbf{Df Residuals:}     &           2      & \textbf{  BIC:               }          &    -1.885   \\
\textbf{Df Model:}         &           4      & \textbf{                     }          &             \\
\bottomrule
\end{tabular}
\begin{tabular}{lcccccc}
            & \textbf{coef} & \textbf{std err} & \textbf{t} & \textbf{P$> |$t$|$} & \textbf{[0.025} & \textbf{0.975]}  \\
\midrule
\textbf{Humans} &       0.0982  &        0.107     &     0.914  &         0.457        &       -0.364    &        0.560     \\
\textbf{ResNet50} &       0.4798  &        0.570     &     0.841  &         0.489        &       -1.974    &        2.933     \\
\textbf{ResNet101} &       0.3946  &        2.177     &     0.181  &         0.873        &       -8.973    &        9.762     \\
\textbf{VOneResNet50} &       3.5181  &        2.264     &     1.554  &         0.260        &       -6.225    &       13.261     \\
\bottomrule
\end{tabular}
\begin{tabular}{lclc}
\textbf{Omnibus:}       &    nan & \textbf{  Durbin-Watson:     } &    2.906  \\
\textbf{Prob(Omnibus):} &    nan & \textbf{  Jarque-Bera (JB):  } &    0.324  \\
\textbf{Skew:}          &  0.467 & \textbf{  Prob(JB):          } &    0.850  \\
\textbf{Kurtosis:}      &  2.348 & \textbf{  Cond. No.          } &     96.8  \\
\bottomrule
\end{tabular}
\end{table}
%\caption{OLS Regression Results}
\end{center}

% running OLS test for  Seg
\begin{center}
\begin{table}
\caption{OLS for all segmentation shuffles}
\label{tab:olssegall}
\begin{tabular}{lclc}
\toprule
\textbf{Dep. Variable:}    &        y         & \textbf{  R-squared (uncentered):}      &     0.959   \\
\textbf{Model:}            &       OLS        & \textbf{  Adj. R-squared (uncentered):} &     0.925   \\
\textbf{Method:}           &  Least Squares   & \textbf{  F-statistic:       }          &     28.91   \\
\textbf{Date:}             & Sat, 26 Nov 2022 & \textbf{  Prob (F-statistic):}          &  0.00119    \\
\textbf{Time:}             &     13:28:32     & \textbf{  Log-Likelihood:    }          &    1.5554   \\
\textbf{No. Observations:} &           9      & \textbf{  AIC:               }          &     4.889   \\
\textbf{Df Residuals:}     &           5      & \textbf{  BIC:               }          &     5.678   \\
\textbf{Df Model:}         &           4      & \textbf{                     }          &             \\
\bottomrule
\end{tabular}
\begin{tabular}{lcccccc}
            & \textbf{coef} & \textbf{std err} & \textbf{t} & \textbf{P$> |$t$|$} & \textbf{[0.025} & \textbf{0.975]}  \\
\midrule
\textbf{Humans} &       0.1604  &        0.062     &     2.575  &         0.050        &        0.000    &        0.320     \\
\textbf{ResNet50} &       0.7505  &        0.695     &     1.080  &         0.330        &       -1.036    &        2.537     \\
\textbf{ResNet101} &      -0.7065  &        2.035     &    -0.347  &         0.743        &       -5.938    &        4.525     \\
\textbf{VOneResNet50} &       2.8102  &        2.297     &     1.224  &         0.276        &       -3.093    &        8.714     \\
\bottomrule
\end{tabular}
\begin{tabular}{lclc}
\textbf{Omnibus:}       &  1.418 & \textbf{  Durbin-Watson:     } &    1.607  \\
\textbf{Prob(Omnibus):} &  0.492 & \textbf{  Jarque-Bera (JB):  } &    0.970  \\
\textbf{Skew:}          &  0.597 & \textbf{  Prob(JB):          } &    0.616  \\
\textbf{Kurtosis:}      &  1.922 & \textbf{  Cond. No.          } &     97.5  \\
\bottomrule
\end{tabular}
\end{table}
%\caption{OLS Regression Results}
\end{center}

\end{appendices}

%%===========================================================================================%%
%% If you are submitting to one of the Nature Portfolio journals, using the eJP submission   %%
%% system, please include the references within the manuscript file itself. You may do this  %%
%% by copying the reference list from your .bbl file, paste it into the main manuscript .tex %%
%% file, and delete the associated \verb+\bibliography+ commands.                            %%
%%===========================================================================================%%
\clearpage
% \bibliography{sn-bibliography}% common bib file
\bibliography{sn-article}% common bib file

%% BioMed_Central_Bib_Style_v1.01

\begin{thebibliography}{91}
% BibTex style file: bmc-mathphys.bst (version 2.1), 2014-07-24
\ifx \bisbn   \undefined \def \bisbn  #1{ISBN #1}\fi
\ifx \binits  \undefined \def \binits#1{#1}\fi
\ifx \bauthor  \undefined \def \bauthor#1{#1}\fi
\ifx \batitle  \undefined \def \batitle#1{#1}\fi
\ifx \bjtitle  \undefined \def \bjtitle#1{#1}\fi
\ifx \bvolume  \undefined \def \bvolume#1{\textbf{#1}}\fi
\ifx \byear  \undefined \def \byear#1{#1}\fi
\ifx \bissue  \undefined \def \bissue#1{#1}\fi
\ifx \bfpage  \undefined \def \bfpage#1{#1}\fi
\ifx \blpage  \undefined \def \blpage #1{#1}\fi
\ifx \burl  \undefined \def \burl#1{\textsf{#1}}\fi
\ifx \doiurl  \undefined \def \doiurl#1{\url{https://doi.org/#1}}\fi
\ifx \betal  \undefined \def \betal{\textit{et al.}}\fi
\ifx \binstitute  \undefined \def \binstitute#1{#1}\fi
\ifx \binstitutionaled  \undefined \def \binstitutionaled#1{#1}\fi
\ifx \bctitle  \undefined \def \bctitle#1{#1}\fi
\ifx \beditor  \undefined \def \beditor#1{#1}\fi
\ifx \bpublisher  \undefined \def \bpublisher#1{#1}\fi
\ifx \bbtitle  \undefined \def \bbtitle#1{#1}\fi
\ifx \bedition  \undefined \def \bedition#1{#1}\fi
\ifx \bseriesno  \undefined \def \bseriesno#1{#1}\fi
\ifx \blocation  \undefined \def \blocation#1{#1}\fi
\ifx \bsertitle  \undefined \def \bsertitle#1{#1}\fi
\ifx \bsnm \undefined \def \bsnm#1{#1}\fi
\ifx \bsuffix \undefined \def \bsuffix#1{#1}\fi
\ifx \bparticle \undefined \def \bparticle#1{#1}\fi
\ifx \barticle \undefined \def \barticle#1{#1}\fi
\bibcommenthead
\ifx \bconfdate \undefined \def \bconfdate #1{#1}\fi
\ifx \botherref \undefined \def \botherref #1{#1}\fi
\ifx \url \undefined \def \url#1{\textsf{#1}}\fi
\ifx \bchapter \undefined \def \bchapter#1{#1}\fi
\ifx \bbook \undefined \def \bbook#1{#1}\fi
\ifx \bcomment \undefined \def \bcomment#1{#1}\fi
\ifx \oauthor \undefined \def \oauthor#1{#1}\fi
\ifx \citeauthoryear \undefined \def \citeauthoryear#1{#1}\fi
\ifx \endbibitem  \undefined \def \endbibitem {}\fi
\ifx \bconflocation  \undefined \def \bconflocation#1{#1}\fi
\ifx \arxivurl  \undefined \def \arxivurl#1{\textsf{#1}}\fi
\csname PreBibitemsHook\endcsname

%%% 1
\bibitem{saisan2001dynamic}
\begin{bchapter}
\bauthor{\bsnm{Saisan}, \binits{P.}},
\bauthor{\bsnm{Doretto}, \binits{G.}},
\bauthor{\bsnm{Wu}, \binits{Y.N.}},
\bauthor{\bsnm{Soatto}, \binits{S.}}:
\bctitle{Dynamic texture recognition}.
In: \bbtitle{Proceedings of the 2001 IEEE Computer Society Conference on
  Computer Vision and Pattern Recognition. CVPR 2001},
vol. \bseriesno{2},
p.
(\byear{2001}).
\bcomment{IEEE}
\end{bchapter}
\endbibitem

%%% 2
\bibitem{renninger2004scene}
\begin{barticle}
\bauthor{\bsnm{Renninger}, \binits{L.W.}},
\bauthor{\bsnm{Malik}, \binits{J.}}:
\batitle{When is scene identification just texture recognition?}
\bjtitle{Vision research}
\bvolume{44}(\bissue{19}),
\bfpage{2301}--\blpage{2311}
(\byear{2004})
\end{barticle}
\endbibitem

%%% 3
\bibitem{kellokumpu2011recognition}
\begin{barticle}
\bauthor{\bsnm{Kellokumpu}, \binits{V.}},
\bauthor{\bsnm{Zhao}, \binits{G.}},
\bauthor{\bsnm{Pietik{\"a}inen}, \binits{M.}}:
\batitle{Recognition of human actions using texture descriptors}.
\bjtitle{Machine Vision and Applications}
\bvolume{22}(\bissue{5}),
\bfpage{767}--\blpage{780}
(\byear{2011})
\end{barticle}
\endbibitem

%%% 4
\bibitem{de1998texture}
\begin{bchapter}
\bauthor{\bsnm{De~Bonet}, \binits{J.S.}},
\bauthor{\bsnm{Viola}, \binits{P.}}:
\bctitle{Texture recognition using a non-parametric multi-scale statistical
  model}.
In: \bbtitle{Proceedings. 1998 IEEE Computer Society Conference on Computer
  Vision and Pattern Recognition (Cat. No. 98CB36231)},
pp. \bfpage{641}--\blpage{647}
(\byear{1998}).
\bcomment{IEEE}
\end{bchapter}
\endbibitem

%%% 5
\bibitem{chaaraoui2013silhouette}
\begin{barticle}
\bauthor{\bsnm{Chaaraoui}, \binits{A.A.}},
\bauthor{\bsnm{Climent-P{\'e}rez}, \binits{P.}},
\bauthor{\bsnm{Fl{\'o}rez-Revuelta}, \binits{F.}}:
\batitle{Silhouette-based human action recognition using sequences of key
  poses}.
\bjtitle{Pattern Recognition Letters}
\bvolume{34}(\bissue{15}),
\bfpage{1799}--\blpage{1807}
(\byear{2013})
\end{barticle}
\endbibitem

%%% 6
\bibitem{al2015human}
\begin{bchapter}
\bauthor{\bsnm{Al-Ali}, \binits{S.}},
\bauthor{\bsnm{Milanova}, \binits{M.}},
\bauthor{\bsnm{Al-Rizzo}, \binits{H.}},
\bauthor{\bsnm{Fox}, \binits{V.L.}}:
\bctitle{Human action recognition: contour-based and silhouette-based
  approaches}.
In: \bbtitle{Computer Vision in Control Systems-2},
pp. \bfpage{11}--\blpage{47}.
\bpublisher{Springer}, \blocation{???}
(\byear{2015})
\end{bchapter}
\endbibitem

%%% 7
\bibitem{popoola2012video}
\begin{barticle}
\bauthor{\bsnm{Popoola}, \binits{O.P.}},
\bauthor{\bsnm{Wang}, \binits{K.}}:
\batitle{Video-based abnormal human behavior recognition—a review}.
\bjtitle{IEEE Transactions on Systems, Man, and Cybernetics, Part C
  (Applications and Reviews)}
\bvolume{42}(\bissue{6}),
\bfpage{865}--\blpage{878}
(\byear{2012})
\end{barticle}
\endbibitem

%%% 8
\bibitem{oliva2007role}
\begin{barticle}
\bauthor{\bsnm{Oliva}, \binits{A.}},
\bauthor{\bsnm{Torralba}, \binits{A.}}:
\batitle{The role of context in object recognition}.
\bjtitle{Trends in cognitive sciences}
\bvolume{11}(\bissue{12}),
\bfpage{520}--\blpage{527}
(\byear{2007})
\end{barticle}
\endbibitem

%%% 9
\bibitem{zhang2020putting}
\begin{bchapter}
\bauthor{\bsnm{Zhang}, \binits{M.}},
\bauthor{\bsnm{Tseng}, \binits{C.}},
\bauthor{\bsnm{Kreiman}, \binits{G.}}:
\bctitle{Putting visual object recognition in context}.
In: \bbtitle{Proceedings of the IEEE/CVF Conference on Computer Vision and
  Pattern Recognition},
pp. \bfpage{12985}--\blpage{12994}
(\byear{2020})
\end{bchapter}
\endbibitem

%%% 10
\bibitem{mori2004recovering}
\begin{bchapter}
\bauthor{\bsnm{Mori}, \binits{G.}},
\bauthor{\bsnm{Ren}, \binits{X.}},
\bauthor{\bsnm{Efros}, \binits{A.A.}},
\bauthor{\bsnm{Malik}, \binits{J.}}:
\bctitle{Recovering human body configurations: Combining segmentation and
  recognition}.
In: \bbtitle{Proceedings of the 2004 IEEE Computer Society Conference on
  Computer Vision and Pattern Recognition, 2004. CVPR 2004.},
vol. \bseriesno{2},
p.
(\byear{2004}).
\bcomment{IEEE}
\end{bchapter}
\endbibitem

%%% 11
\bibitem{beleznai2009fast}
\begin{bchapter}
\bauthor{\bsnm{Beleznai}, \binits{C.}},
\bauthor{\bsnm{Bischof}, \binits{H.}}:
\bctitle{Fast human detection in crowded scenes by contour integration and
  local shape estimation}.
In: \bbtitle{2009 IEEE Conference on Computer Vision and Pattern Recognition},
pp. \bfpage{2246}--\blpage{2253}
(\byear{2009}).
\bcomment{IEEE}
\end{bchapter}
\endbibitem

%%% 12
\bibitem{zhou2019humans}
\begin{barticle}
\bauthor{\bsnm{Zhou}, \binits{Z.}},
\bauthor{\bsnm{Firestone}, \binits{C.}}:
\batitle{Humans can decipher adversarial images}.
\bjtitle{Nature communications}
\bvolume{10}(\bissue{1}),
\bfpage{1}--\blpage{9}
(\byear{2019})
\end{barticle}
\endbibitem

%%% 13
\bibitem{eidolons}
\begin{barticle}
\bauthor{\bsnm{Koenderink}, \binits{J.}},
\bauthor{\bsnm{Valsecchi}, \binits{M.}},
\bauthor{\bparticle{van} \bsnm{Doorn}, \binits{A.}},
\bauthor{\bsnm{Wagemans}, \binits{J.}},
\bauthor{\bsnm{Gegenfurtner}, \binits{K.}}:
\batitle{Eidolons: Novel stimuli for vision research}.
\bjtitle{Journal of Vision}
\bvolume{17}(\bissue{2}),
\bfpage{7}--\blpage{7}
(\byear{2017})
\end{barticle}
\endbibitem

%%% 14
\bibitem{geirhos2018generalisation}
\begin{botherref}
\oauthor{\bsnm{Geirhos}, \binits{R.}},
\oauthor{\bsnm{Temme}, \binits{C.R.}},
\oauthor{\bsnm{Rauber}, \binits{J.}},
\oauthor{\bsnm{Sch{\"u}tt}, \binits{H.H.}},
\oauthor{\bsnm{Bethge}, \binits{M.}},
\oauthor{\bsnm{Wichmann}, \binits{F.A.}}:
Generalisation in humans and deep neural networks.
Advances in neural information processing systems
\textbf{31}
(2018)
\end{botherref}
\endbibitem

%%% 15
\bibitem{geirhos2020shortcut}
\begin{barticle}
\bauthor{\bsnm{Geirhos}, \binits{R.}},
\bauthor{\bsnm{Jacobsen}, \binits{J.-H.}},
\bauthor{\bsnm{Michaelis}, \binits{C.}},
\bauthor{\bsnm{Zemel}, \binits{R.}},
\bauthor{\bsnm{Brendel}, \binits{W.}},
\bauthor{\bsnm{Bethge}, \binits{M.}},
\bauthor{\bsnm{Wichmann}, \binits{F.A.}}:
\batitle{Shortcut learning in deep neural networks}.
\bjtitle{Nature Machine Intelligence}
\bvolume{2}(\bissue{11}),
\bfpage{665}--\blpage{673}
(\byear{2020})
\end{barticle}
\endbibitem

%%% 16
\bibitem{yamins2016using}
\begin{barticle}
\bauthor{\bsnm{Yamins}, \binits{D.L.}},
\bauthor{\bsnm{DiCarlo}, \binits{J.J.}}:
\batitle{Using goal-driven deep learning models to understand sensory cortex}.
\bjtitle{Nature neuroscience}
\bvolume{19}(\bissue{3}),
\bfpage{356}--\blpage{365}
(\byear{2016})
\end{barticle}
\endbibitem

%%% 17
\bibitem{dong2018commentary}
\begin{barticle}
\bauthor{\bsnm{Dong}, \binits{Q.}},
\bauthor{\bsnm{Wang}, \binits{H.}},
\bauthor{\bsnm{Hu}, \binits{Z.}}:
\batitle{Commentary: Using goal-driven deep learning models to understand
  sensory cortex}.
\bjtitle{Frontiers in Computational Neuroscience}
\bvolume{12},
\bfpage{4}
(\byear{2018})
\end{barticle}
\endbibitem

%%% 18
\bibitem{bear2020neuroscience}
\begin{bbook}
\bauthor{\bsnm{Bear}, \binits{M.}},
\bauthor{\bsnm{Connors}, \binits{B.}},
\bauthor{\bsnm{Paradiso}, \binits{M.A.}}:
\bbtitle{Neuroscience: Exploring the Brain, Enhanced Edition: Exploring the
  Brain, Enhanced Edition}.
\bpublisher{Jones \& Bartlett Learning}, \blocation{???}
(\byear{2020}).
\burl{https://books.google.com/books?id=m-PcDwAAQBAJ}
\end{bbook}
\endbibitem

%%% 19
\bibitem{mumford1}
\begin{barticle}
\bauthor{\bsnm{Lee}, \binits{T.S.}},
\bauthor{\bsnm{Mumford}, \binits{D.}}:
\batitle{Hierarchical bayesian inference in the visual cortex}.
\bjtitle{JOSA A}
\bvolume{20}(\bissue{7}),
\bfpage{1434}--\blpage{1448}
(\byear{2003})
\end{barticle}
\endbibitem

%%% 20
\bibitem{mumford2}
\begin{barticle}
\bauthor{\bsnm{Zhu}, \binits{S.-C.}},
\bauthor{\bsnm{Mumford}, \binits{D.}}, \betal:
\batitle{A stochastic grammar of images}.
\bjtitle{Foundations and Trends{\textregistered} in Computer Graphics and
  Vision}
\bvolume{2}(\bissue{4}),
\bfpage{259}--\blpage{362}
(\byear{2007})
\end{barticle}
\endbibitem

%%% 21
\bibitem{hochstein2002view}
\begin{barticle}
\bauthor{\bsnm{Hochstein}, \binits{S.}},
\bauthor{\bsnm{Ahissar}, \binits{M.}}:
\batitle{View from the top: Hierarchies and reverse hierarchies in the visual
  system}.
\bjtitle{Neuron}
\bvolume{36}(\bissue{5}),
\bfpage{791}--\blpage{804}
(\byear{2002})
\end{barticle}
\endbibitem

%%% 22
\bibitem{corbett2023pervasiveness}
\begin{bbook}
\bauthor{\bsnm{Corbett}, \binits{J.E.}},
\bauthor{\bsnm{Utochkin}, \binits{I.}},
\bauthor{\bsnm{Hochstein}, \binits{S.}}:
\bbtitle{The Pervasiveness of Ensemble Perception: Not Just Your Average
  Review}.
\bpublisher{Cambridge University Press}, \blocation{???}
(\byear{2023})
\end{bbook}
\endbibitem

%%% 23
\bibitem{ullman2016atoms}
\begin{barticle}
\bauthor{\bsnm{Ullman}, \binits{S.}},
\bauthor{\bsnm{Assif}, \binits{L.}},
\bauthor{\bsnm{Fetaya}, \binits{E.}},
\bauthor{\bsnm{Harari}, \binits{D.}}:
\batitle{Atoms of recognition in human and computer vision}.
\bjtitle{Proceedings of the National Academy of Sciences}
\bvolume{113}(\bissue{10}),
\bfpage{2744}--\blpage{2749}
(\byear{2016})
\end{barticle}
\endbibitem

%%% 24
\bibitem{edelman1997complex}
\begin{botherref}
\oauthor{\bsnm{Edelman}, \binits{S.}},
\oauthor{\bsnm{Intrator}, \binits{N.}},
\oauthor{\bsnm{Poggio}, \binits{T.}}:
Complex cells and object recognition
(1997)
\end{botherref}
\endbibitem

%%% 25
\bibitem{tarr1998image}
\begin{barticle}
\bauthor{\bsnm{Tarr}, \binits{M.J.}},
\bauthor{\bsnm{B{\"u}lthoff}, \binits{H.H.}}:
\batitle{Image-based object recognition in man, monkey and machine}.
\bjtitle{Cognition}
\bvolume{67}(\bissue{1-2}),
\bfpage{1}--\blpage{20}
(\byear{1998})
\end{barticle}
\endbibitem

%%% 26
\bibitem{grill2001lateral}
\begin{barticle}
\bauthor{\bsnm{Grill-Spector}, \binits{K.}},
\bauthor{\bsnm{Kourtzi}, \binits{Z.}},
\bauthor{\bsnm{Kanwisher}, \binits{N.}}:
\batitle{The lateral occipital complex and its role in object recognition}.
\bjtitle{Vision research}
\bvolume{41}(\bissue{10-11}),
\bfpage{1409}--\blpage{1422}
(\byear{2001})
\end{barticle}
\endbibitem

%%% 27
\bibitem{biederman1991priming}
\begin{barticle}
\bauthor{\bsnm{Biederman}, \binits{I.}},
\bauthor{\bsnm{Cooper}, \binits{E.E.}}:
\batitle{Priming contour-deleted images: Evidence for intermediate
  representations in visual object recognition}.
\bjtitle{Cognitive psychology}
\bvolume{23}(\bissue{3}),
\bfpage{393}--\blpage{419}
(\byear{1991})
\end{barticle}
\endbibitem

%%% 28
\bibitem{ferrari2007groups}
\begin{barticle}
\bauthor{\bsnm{Ferrari}, \binits{V.}},
\bauthor{\bsnm{Fevrier}, \binits{L.}},
\bauthor{\bsnm{Jurie}, \binits{F.}},
\bauthor{\bsnm{Schmid}, \binits{C.}}:
\batitle{Groups of adjacent contour segments for object detection}.
\bjtitle{IEEE transactions on pattern analysis and machine intelligence}
\bvolume{30}(\bissue{1}),
\bfpage{36}--\blpage{51}
(\byear{2007})
\end{barticle}
\endbibitem

%%% 29
\bibitem{rusak2020simple}
\begin{bchapter}
\bauthor{\bsnm{Rusak}, \binits{E.}},
\bauthor{\bsnm{Schott}, \binits{L.}},
\bauthor{\bsnm{Zimmermann}, \binits{R.S.}},
\bauthor{\bsnm{Bitterwolf}, \binits{J.}},
\bauthor{\bsnm{Bringmann}, \binits{O.}},
\bauthor{\bsnm{Bethge}, \binits{M.}},
\bauthor{\bsnm{Brendel}, \binits{W.}}:
\bctitle{A simple way to make neural networks robust against diverse image
  corruptions}.
In: \bbtitle{European Conference on Computer Vision},
pp. \bfpage{53}--\blpage{69}
(\byear{2020}).
\bcomment{Springer}
\end{bchapter}
\endbibitem

%%% 30
\bibitem{baker2018deep}
\begin{barticle}
\bauthor{\bsnm{Baker}, \binits{N.}},
\bauthor{\bsnm{Lu}, \binits{H.}},
\bauthor{\bsnm{Erlikhman}, \binits{G.}},
\bauthor{\bsnm{Kellman}, \binits{P.J.}}:
\batitle{Deep convolutional networks do not classify based on global object
  shape}.
\bjtitle{PLoS computational biology}
\bvolume{14}(\bissue{12}),
\bfpage{1006613}
(\byear{2018})
\end{barticle}
\endbibitem

%%% 31
\bibitem{learningtoseebylookingatnoise}
\begin{barticle}
\bauthor{\bsnm{Baradad~Jurjo}, \binits{M.}},
\bauthor{\bsnm{Wulff}, \binits{J.}},
\bauthor{\bsnm{Wang}, \binits{T.}},
\bauthor{\bsnm{Isola}, \binits{P.}},
\bauthor{\bsnm{Torralba}, \binits{A.}}:
\batitle{Learning to see by looking at noise}.
\bjtitle{Advances in Neural Information Processing Systems}
\bvolume{34},
\bfpage{2556}--\blpage{2569}
(\byear{2021})
\end{barticle}
\endbibitem

%%% 32
\bibitem{nguyen2015deep}
\begin{bchapter}
\bauthor{\bsnm{Nguyen}, \binits{A.}},
\bauthor{\bsnm{Yosinski}, \binits{J.}},
\bauthor{\bsnm{Clune}, \binits{J.}}:
\bctitle{Deep neural networks are easily fooled: High confidence predictions
  for unrecognizable images}.
In: \bbtitle{Proceedings of the IEEE Conference on Computer Vision and Pattern
  Recognition},
pp. \bfpage{427}--\blpage{436}
(\byear{2015})
\end{bchapter}
\endbibitem

%%% 33
\bibitem{imagenet}
\begin{bchapter}
\bauthor{\bsnm{Deng}, \binits{J.}},
\bauthor{\bsnm{Dong}, \binits{W.}},
\bauthor{\bsnm{Socher}, \binits{R.}},
\bauthor{\bsnm{Li}, \binits{L.-J.}},
\bauthor{\bsnm{Li}, \binits{K.}},
\bauthor{\bsnm{Fei-Fei}, \binits{L.}}:
\bctitle{Imagenet: A large-scale hierarchical image database}.
In: \bbtitle{2009 IEEE Conference on Computer Vision and Pattern Recognition},
pp. \bfpage{248}--\blpage{255}
(\byear{2009}).
\bcomment{IEEE}
\end{bchapter}
\endbibitem

%%% 34
\bibitem{geirhos2018imagenettrained}
\begin{bchapter}
\bauthor{\bsnm{Geirhos}, \binits{R.}},
\bauthor{\bsnm{Rubisch}, \binits{P.}},
\bauthor{\bsnm{Michaelis}, \binits{C.}},
\bauthor{\bsnm{Bethge}, \binits{M.}},
\bauthor{\bsnm{Wichmann}, \binits{F.A.}},
\bauthor{\bsnm{Brendel}, \binits{W.}}:
\bctitle{Imagenet-trained {CNN}s are biased towards texture; increasing shape
  bias improves accuracy and robustness.}
In: \bbtitle{International Conference on Learning Representations}
(\byear{2019}).
\burl{https://openreview.net/forum?id=Bygh9j09KX}
\end{bchapter}
\endbibitem

%%% 35
\bibitem{resnet}
\begin{bchapter}
\bauthor{\bsnm{He}, \binits{K.}},
\bauthor{\bsnm{Zhang}, \binits{X.}},
\bauthor{\bsnm{Ren}, \binits{S.}},
\bauthor{\bsnm{Sun}, \binits{J.}}:
\bctitle{Deep residual learning for image recognition}.
In: \bbtitle{Proceedings of the IEEE Conference on Computer Vision and Pattern
  Recognition},
pp. \bfpage{770}--\blpage{778}
(\byear{2016})
\end{bchapter}
\endbibitem

%%% 36
\bibitem{gatys2017texture}
\begin{barticle}
\bauthor{\bsnm{Gatys}, \binits{L.A.}},
\bauthor{\bsnm{Ecker}, \binits{A.S.}},
\bauthor{\bsnm{Bethge}, \binits{M.}}:
\batitle{Texture and art with deep neural networks}.
\bjtitle{Current opinion in neurobiology}
\bvolume{46},
\bfpage{178}--\blpage{186}
(\byear{2017})
\end{barticle}
\endbibitem

%%% 37
\bibitem{brendel2019approximating}
\begin{botherref}
\oauthor{\bsnm{Brendel}, \binits{W.}},
\oauthor{\bsnm{Bethge}, \binits{M.}}:
Approximating cnns with bag-of-local-features models works surprisingly well on
  imagenet.
arXiv preprint arXiv:1904.00760
(2019)
\end{botherref}
\endbibitem

%%% 38
\bibitem{yu2017sketch}
\begin{barticle}
\bauthor{\bsnm{Yu}, \binits{Q.}},
\bauthor{\bsnm{Yang}, \binits{Y.}},
\bauthor{\bsnm{Liu}, \binits{F.}},
\bauthor{\bsnm{Song}, \binits{Y.-Z.}},
\bauthor{\bsnm{Xiang}, \binits{T.}},
\bauthor{\bsnm{Hospedales}, \binits{T.M.}}:
\batitle{Sketch-a-net: A deep neural network that beats humans}.
\bjtitle{International journal of computer vision}
\bvolume{122},
\bfpage{411}--\blpage{425}
(\byear{2017})
\end{barticle}
\endbibitem

%%% 39
\bibitem{ballester2016performance}
\begin{bchapter}
\bauthor{\bsnm{Ballester}, \binits{P.}},
\bauthor{\bsnm{Araujo}, \binits{R.}}:
\bctitle{On the performance of googlenet and alexnet applied to sketches}.
In: \bbtitle{Proceedings of the AAAI Conference on Artificial Intelligence},
vol. \bseriesno{30}
(\byear{2016})
\end{bchapter}
\endbibitem

%%% 40
\bibitem{tolstikhin2021mlpmixer}
\begin{bchapter}
\bauthor{\bsnm{Tolstikhin}, \binits{I.}},
\bauthor{\bsnm{Houlsby}, \binits{N.}},
\bauthor{\bsnm{Kolesnikov}, \binits{A.}},
\bauthor{\bsnm{Beyer}, \binits{L.}},
\bauthor{\bsnm{Zhai}, \binits{X.}},
\bauthor{\bsnm{Unterthiner}, \binits{T.}},
\bauthor{\bsnm{Yung}, \binits{J.}},
\bauthor{\bsnm{Steiner}, \binits{A.P.}},
\bauthor{\bsnm{Keysers}, \binits{D.}},
\bauthor{\bsnm{Uszkoreit}, \binits{J.}},
\bauthor{\bsnm{Lucic}, \binits{M.}},
\bauthor{\bsnm{Dosovitskiy}, \binits{A.}}:
\bctitle{{MLP}-mixer: An all-{MLP} architecture for vision}.
In: \beditor{\bsnm{Beygelzimer}, \binits{A.}},
\beditor{\bsnm{Dauphin}, \binits{Y.}},
\beditor{\bsnm{Liang}, \binits{P.}},
\beditor{\bsnm{Vaughan}, \binits{J.W.}} (eds.)
\bbtitle{Advances in Neural Information Processing Systems}
(\byear{2021}).
\burl{https://openreview.net/forum?id=EI2KOXKdnP}
\end{bchapter}
\endbibitem

%%% 41
\bibitem{dosovitskiy2021an}
\begin{bchapter}
\bauthor{\bsnm{Dosovitskiy}, \binits{A.}},
\bauthor{\bsnm{Beyer}, \binits{L.}},
\bauthor{\bsnm{Kolesnikov}, \binits{A.}},
\bauthor{\bsnm{Weissenborn}, \binits{D.}},
\bauthor{\bsnm{Zhai}, \binits{X.}},
\bauthor{\bsnm{Unterthiner}, \binits{T.}},
\bauthor{\bsnm{Dehghani}, \binits{M.}},
\bauthor{\bsnm{Minderer}, \binits{M.}},
\bauthor{\bsnm{Heigold}, \binits{G.}},
\bauthor{\bsnm{Gelly}, \binits{S.}},
\bauthor{\bsnm{Uszkoreit}, \binits{J.}},
\bauthor{\bsnm{Houlsby}, \binits{N.}}:
\bctitle{An image is worth 16x16 words: Transformers for image recognition at
  scale}.
In: \bbtitle{International Conference on Learning Representations}
(\byear{2021}).
\burl{https://openreview.net/forum?id=YicbFdNTTy}
\end{bchapter}
\endbibitem

%%% 42
\bibitem{dapello2021neural}
\begin{barticle}
\bauthor{\bsnm{Dapello}, \binits{J.}},
\bauthor{\bsnm{Feather}, \binits{J.}},
\bauthor{\bsnm{Le}, \binits{H.}},
\bauthor{\bsnm{Marques}, \binits{T.}},
\bauthor{\bsnm{Cox}, \binits{D.}},
\bauthor{\bsnm{McDermott}, \binits{J.}},
\bauthor{\bsnm{DiCarlo}, \binits{J.J.}},
\bauthor{\bsnm{Chung}, \binits{S.}}:
\batitle{Neural population geometry reveals the role of stochasticity in robust
  perception}.
\bjtitle{Advances in Neural Information Processing Systems}
\bvolume{34},
\bfpage{15595}--\blpage{15607}
(\byear{2021})
\end{barticle}
\endbibitem

%%% 43
\bibitem{crowder2022robustness}
\begin{botherref}
\oauthor{\bsnm{Crowder}, \binits{D.}},
\oauthor{\bsnm{Malik}, \binits{G.}}:
Robustness of humans and machines on object recognition with extreme image
  transformations.
CVPR Workshop on What can computer vision learn from visual neuroscience?
(2022)
\end{botherref}
\endbibitem

%%% 44
\bibitem{he2015delving}
\begin{bchapter}
\bauthor{\bsnm{He}, \binits{K.}},
\bauthor{\bsnm{Zhang}, \binits{X.}},
\bauthor{\bsnm{Ren}, \binits{S.}},
\bauthor{\bsnm{Sun}, \binits{J.}}:
\bctitle{Delving deep into rectifiers: Surpassing human-level performance on
  imagenet classification}.
In: \bbtitle{Proceedings of the IEEE International Conference on Computer
  Vision},
pp. \bfpage{1026}--\blpage{1034}
(\byear{2015})
\end{bchapter}
\endbibitem

%%% 45
\bibitem{taigman2014deepface}
\begin{bchapter}
\bauthor{\bsnm{Taigman}, \binits{Y.}},
\bauthor{\bsnm{Yang}, \binits{M.}},
\bauthor{\bsnm{Ranzato}, \binits{M.}},
\bauthor{\bsnm{Wolf}, \binits{L.}}:
\bctitle{Deepface: Closing the gap to human-level performance in face
  verification}.
In: \bbtitle{Proceedings of the IEEE Conference on Computer Vision and Pattern
  Recognition},
pp. \bfpage{1701}--\blpage{1708}
(\byear{2014})
\end{bchapter}
\endbibitem

%%% 46
\bibitem{mnih2015human}
\begin{barticle}
\bauthor{\bsnm{Mnih}, \binits{V.}},
\bauthor{\bsnm{Kavukcuoglu}, \binits{K.}},
\bauthor{\bsnm{Silver}, \binits{D.}},
\bauthor{\bsnm{Rusu}, \binits{A.A.}},
\bauthor{\bsnm{Veness}, \binits{J.}},
\bauthor{\bsnm{Bellemare}, \binits{M.G.}},
\bauthor{\bsnm{Graves}, \binits{A.}},
\bauthor{\bsnm{Riedmiller}, \binits{M.}},
\bauthor{\bsnm{Fidjeland}, \binits{A.K.}},
\bauthor{\bsnm{Ostrovski}, \binits{G.}}, \betal:
\batitle{Human-level control through deep reinforcement learning}.
\bjtitle{nature}
\bvolume{518}(\bissue{7540}),
\bfpage{529}--\blpage{533}
(\byear{2015})
\end{barticle}
\endbibitem

%%% 47
\bibitem{douglas1991functional}
\begin{barticle}
\bauthor{\bsnm{Douglas}, \binits{R.J.}},
\bauthor{\bsnm{Martin}, \binits{K.}}:
\batitle{A functional microcircuit for cat visual cortex.}
\bjtitle{The Journal of physiology}
\bvolume{440}(\bissue{1}),
\bfpage{735}--\blpage{769}
(\byear{1991})
\end{barticle}
\endbibitem

%%% 48
\bibitem{wilson1991computer}
\begin{barticle}
\bauthor{\bsnm{Wilson}, \binits{M.A.}},
\bauthor{\bsnm{Bower}, \binits{J.M.}}:
\batitle{A computer simulation of oscillatory behavior in primary visual
  cortex}.
\bjtitle{Neural Computation}
\bvolume{3}(\bissue{4}),
\bfpage{498}--\blpage{509}
(\byear{1991})
\end{barticle}
\endbibitem

%%% 49
\bibitem{bednar2012building}
\begin{barticle}
\bauthor{\bsnm{Bednar}, \binits{J.A.}}:
\batitle{Building a mechanistic model of the development and function of the
  primary visual cortex}.
\bjtitle{Journal of Physiology-Paris}
\bvolume{106}(\bissue{5-6}),
\bfpage{194}--\blpage{211}
(\byear{2012})
\end{barticle}
\endbibitem

%%% 50
\bibitem{wang2008perceptual}
\begin{barticle}
\bauthor{\bsnm{Wang}, \binits{Y.}},
\bauthor{\bsnm{Zhu}, \binits{S.-C.}}:
\batitle{Perceptual scale-space and its applications}.
\bjtitle{International Journal of Computer Vision}
\bvolume{80},
\bfpage{143}--\blpage{165}
(\byear{2008})
\end{barticle}
\endbibitem

%%% 51
\bibitem{georgeson2007filters}
\begin{barticle}
\bauthor{\bsnm{Georgeson}, \binits{M.A.}},
\bauthor{\bsnm{May}, \binits{K.A.}},
\bauthor{\bsnm{Freeman}, \binits{T.C.}},
\bauthor{\bsnm{Hesse}, \binits{G.S.}}:
\batitle{From filters to features: Scale--space analysis of edge and blur
  coding in human vision}.
\bjtitle{Journal of vision}
\bvolume{7}(\bissue{13}),
\bfpage{7}--\blpage{7}
(\byear{2007})
\end{barticle}
\endbibitem

%%% 52
\bibitem{witkin1987scale}
\begin{bchapter}
\bauthor{\bsnm{Witkin}, \binits{A.P.}}:
\bctitle{Scale-space filtering}.
In: \bbtitle{Readings in Computer Vision},
pp. \bfpage{329}--\blpage{332}.
\bpublisher{Elsevier}, \blocation{???}
(\byear{1987})
\end{bchapter}
\endbibitem

%%% 53
\bibitem{lindeberg2013scale}
\begin{bbook}
\bauthor{\bsnm{Lindeberg}, \binits{T.}}:
\bbtitle{Scale-space Theory in Computer Vision}
vol. \bseriesno{256}.
\bpublisher{Springer}, \blocation{???}
(\byear{2013})
\end{bbook}
\endbibitem

%%% 54
\bibitem{ekstrom2017human}
\begin{barticle}
\bauthor{\bsnm{Ekstrom}, \binits{A.D.}},
\bauthor{\bsnm{Isham}, \binits{E.A.}}:
\batitle{Human spatial navigation: Representations across dimensions and
  scales}.
\bjtitle{Current opinion in behavioral sciences}
\bvolume{17},
\bfpage{84}--\blpage{89}
(\byear{2017})
\end{barticle}
\endbibitem

%%% 55
\bibitem{achanta2010slic}
\begin{botherref}
\oauthor{\bsnm{Achanta}, \binits{R.}},
\oauthor{\bsnm{Shaji}, \binits{A.}},
\oauthor{\bsnm{Smith}, \binits{K.}},
\oauthor{\bsnm{Lucchi}, \binits{A.}},
\oauthor{\bsnm{Fua}, \binits{P.}},
\oauthor{\bsnm{S{\"u}sstrunk}, \binits{S.}}:
Slic superpixels.
Technical report
(2010)
\end{botherref}
\endbibitem

%%% 56
\bibitem{achanta2012slic}
\begin{barticle}
\bauthor{\bsnm{Achanta}, \binits{R.}},
\bauthor{\bsnm{Shaji}, \binits{A.}},
\bauthor{\bsnm{Smith}, \binits{K.}},
\bauthor{\bsnm{Lucchi}, \binits{A.}},
\bauthor{\bsnm{Fua}, \binits{P.}},
\bauthor{\bsnm{S{\"u}sstrunk}, \binits{S.}}:
\batitle{Slic superpixels compared to state-of-the-art superpixel methods}.
\bjtitle{IEEE transactions on pattern analysis and machine intelligence}
\bvolume{34}(\bissue{11}),
\bfpage{2274}--\blpage{2282}
(\byear{2012})
\end{barticle}
\endbibitem

%%% 57
\bibitem{vonenet}
\begin{barticle}
\bauthor{\bsnm{Dapello}, \binits{J.}},
\bauthor{\bsnm{Marques}, \binits{T.}},
\bauthor{\bsnm{Schrimpf}, \binits{M.}},
\bauthor{\bsnm{Geiger}, \binits{F.}},
\bauthor{\bsnm{Cox}, \binits{D.}},
\bauthor{\bsnm{DiCarlo}, \binits{J.J.}}:
\batitle{Simulating a primary visual cortex at the front of cnns improves
  robustness to image perturbations}.
\bjtitle{Advances in Neural Information Processing Systems}
\bvolume{33},
\bfpage{13073}--\blpage{13087}
(\byear{2020})
\end{barticle}
\endbibitem

%%% 58
\bibitem{brainscore}
\begin{botherref}
\oauthor{\bsnm{Schrimpf}, \binits{M.}},
\oauthor{\bsnm{Kubilius}, \binits{J.}},
\oauthor{\bsnm{Hong}, \binits{H.}},
\oauthor{\bsnm{Majaj}, \binits{N.J.}},
\oauthor{\bsnm{Rajalingham}, \binits{R.}},
\oauthor{\bsnm{Issa}, \binits{E.B.}},
\oauthor{\bsnm{Kar}, \binits{K.}},
\oauthor{\bsnm{Bashivan}, \binits{P.}},
\oauthor{\bsnm{Prescott-Roy}, \binits{J.}},
\oauthor{\bsnm{Geiger}, \binits{F.}}, et al.:
Brain-score: Which artificial neural network for object recognition is most
  brain-like?
BioRxiv,
407007
(2020)
\end{botherref}
\endbibitem

%%% 59
\bibitem{imagenettegit}
\begin{botherref}
\oauthor{\bsnm{fast.ai}},
\oauthor{\bsnm{Howard}, \binits{J.}}:
{Imagenette}.
\url{https://github.com/fastai/imagenette}
\end{botherref}
\endbibitem

%%% 60
\bibitem{hubel1962receptive}
\begin{barticle}
\bauthor{\bsnm{Hubel}, \binits{D.H.}},
\bauthor{\bsnm{Wiesel}, \binits{T.N.}}:
\batitle{Receptive fields, binocular interaction and functional architecture in
  the cat's visual cortex}.
\bjtitle{The Journal of physiology}
\bvolume{160}(\bissue{1}),
\bfpage{106}
(\byear{1962})
\end{barticle}
\endbibitem

%%% 61
\bibitem{hubel1963shape}
\begin{barticle}
\bauthor{\bsnm{Hubel}, \binits{D.H.}},
\bauthor{\bsnm{Wiesel}, \binits{T.N.}}:
\batitle{Shape and arrangement of columns in cat's striate cortex}.
\bjtitle{The Journal of physiology}
\bvolume{165}(\bissue{3}),
\bfpage{559}
(\byear{1963})
\end{barticle}
\endbibitem

%%% 62
\bibitem{wiesel1963single}
\begin{barticle}
\bauthor{\bsnm{Wiesel}, \binits{T.N.}},
\bauthor{\bsnm{Hubel}, \binits{D.H.}}:
\batitle{Single-cell responses in striate cortex of kittens deprived of vision
  in one eye}.
\bjtitle{Journal of neurophysiology}
\bvolume{26}(\bissue{6}),
\bfpage{1003}--\blpage{1017}
(\byear{1963})
\end{barticle}
\endbibitem

%%% 63
\bibitem{hubel1963receptive}
\begin{barticle}
\bauthor{\bsnm{Hubel}, \binits{D.H.}},
\bauthor{\bsnm{Wiesel}, \binits{T.N.}}:
\batitle{Receptive fields of cells in striate cortex of very young, visually
  inexperienced kittens}.
\bjtitle{Journal of neurophysiology}
\bvolume{26}(\bissue{6}),
\bfpage{994}--\blpage{1002}
(\byear{1963})
\end{barticle}
\endbibitem

%%% 64
\bibitem{wiesel1963effects}
\begin{barticle}
\bauthor{\bsnm{Wiesel}, \binits{T.N.}},
\bauthor{\bsnm{Hubel}, \binits{D.H.}}:
\batitle{Effects of visual deprivation on morphology and physiology of cells in
  the cat's lateral geniculate body}.
\bjtitle{Journal of neurophysiology}
\bvolume{26}(\bissue{6}),
\bfpage{978}--\blpage{993}
(\byear{1963})
\end{barticle}
\endbibitem

%%% 65
\bibitem{tanaka1997mechanisms}
\begin{barticle}
\bauthor{\bsnm{Tanaka}, \binits{K.}}:
\batitle{Mechanisms of visual object recognition: monkey and human studies}.
\bjtitle{Current opinion in neurobiology}
\bvolume{7}(\bissue{4}),
\bfpage{523}--\blpage{529}
(\byear{1997})
\end{barticle}
\endbibitem

%%% 66
\bibitem{gal2016dropout}
\begin{bchapter}
\bauthor{\bsnm{Gal}, \binits{Y.}},
\bauthor{\bsnm{Ghahramani}, \binits{Z.}}:
\bctitle{Dropout as a bayesian approximation: Representing model uncertainty in
  deep learning}.
In: \bbtitle{International Conference on Machine Learning},
pp. \bfpage{1050}--\blpage{1059}
(\byear{2016}).
\bcomment{PMLR}
\end{bchapter}
\endbibitem

%%% 67
\bibitem{NEURIPS2021linsley}
\begin{bchapter}
\bauthor{\bsnm{Linsley}, \binits{D.}},
\bauthor{\bsnm{Malik}, \binits{G.}},
\bauthor{\bsnm{Kim}, \binits{J.}},
\bauthor{\bsnm{Govindarajan}, \binits{L.N.}},
\bauthor{\bsnm{Mingolla}, \binits{E.}},
\bauthor{\bsnm{Serre}, \binits{T.}}:
\bctitle{Tracking without re-recognition in humans and machines}.
In: \beditor{\bsnm{Ranzato}, \binits{M.}},
\beditor{\bsnm{Beygelzimer}, \binits{A.}},
\beditor{\bsnm{Dauphin}, \binits{Y.}},
\beditor{\bsnm{Liang}, \binits{P.S.}},
\beditor{\bsnm{Vaughan}, \binits{J.W.}} (eds.)
\bbtitle{Advances in Neural Information Processing Systems},
vol. \bseriesno{34},
pp. \bfpage{19473}--\blpage{19486}.
\bpublisher{Curran Associates, Inc.}, \blocation{???}
(\byear{2021}).
\burl{https://proceedings.neurips.cc/paper/2021/file/a2557a7b2e94197ff767970b67041697-Paper.pdf}
\end{bchapter}
\endbibitem

%%% 68
\bibitem{pathtracker}
\begin{botherref}
\oauthor{\bsnm{Malik}, \binits{G.}},
\oauthor{\bsnm{Linsley}, \binits{D.}},
\oauthor{\bsnm{Serre}, \binits{T.}},
\oauthor{\bsnm{Mingolla}, \binits{E.}}:
The challenge of appearance-free object tracking with feedforward neural
  networks.
CVPR Workshop on Dynamic Neural Networks Meet Computer Vision
(2021)
\end{botherref}
\endbibitem

%%% 69
\bibitem{carandini2005we}
\begin{barticle}
\bauthor{\bsnm{Carandini}, \binits{M.}},
\bauthor{\bsnm{Demb}, \binits{J.B.}},
\bauthor{\bsnm{Mante}, \binits{V.}},
\bauthor{\bsnm{Tolhurst}, \binits{D.J.}},
\bauthor{\bsnm{Dan}, \binits{Y.}},
\bauthor{\bsnm{Olshausen}, \binits{B.A.}},
\bauthor{\bsnm{Gallant}, \binits{J.L.}},
\bauthor{\bsnm{Rust}, \binits{N.C.}}:
\batitle{Do we know what the early visual system does?}
\bjtitle{Journal of Neuroscience}
\bvolume{25}(\bissue{46}),
\bfpage{10577}--\blpage{10597}
(\byear{2005})
\end{barticle}
\endbibitem

%%% 70
\bibitem{allison1994human}
\begin{barticle}
\bauthor{\bsnm{Allison}, \binits{T.}},
\bauthor{\bsnm{McCarthy}, \binits{G.}},
\bauthor{\bsnm{Nobre}, \binits{A.}},
\bauthor{\bsnm{Puce}, \binits{A.}},
\bauthor{\bsnm{Belger}, \binits{A.}}:
\batitle{Human extrastriate visual cortex and the perception of faces, words,
  numbers, and colors}.
\bjtitle{Cerebral cortex}
\bvolume{4}(\bissue{5}),
\bfpage{544}--\blpage{554}
(\byear{1994})
\end{barticle}
\endbibitem

%%% 71
\bibitem{martin2016grapes}
\begin{barticle}
\bauthor{\bsnm{Martin}, \binits{A.}}:
\batitle{Grapes—grounding representations in action, perception, and emotion
  systems: How object properties and categories are represented in the human
  brain}.
\bjtitle{Psychonomic bulletin \& review}
\bvolume{23},
\bfpage{979}--\blpage{990}
(\byear{2016})
\end{barticle}
\endbibitem

%%% 72
\bibitem{keil2010feature}
\begin{barticle}
\bauthor{\bsnm{Keil}, \binits{A.}},
\bauthor{\bsnm{M{\"u}ller}, \binits{M.M.}}:
\batitle{Feature selection in the human brain: Electrophysiological correlates
  of sensory enhancement and feature integration}.
\bjtitle{Brain Research}
\bvolume{1313},
\bfpage{172}--\blpage{184}
(\byear{2010})
\end{barticle}
\endbibitem

%%% 73
\bibitem{moon2021integralaction}
\begin{bchapter}
\bauthor{\bsnm{Moon}, \binits{G.}},
\bauthor{\bsnm{Kwon}, \binits{H.}},
\bauthor{\bsnm{Lee}, \binits{K.M.}},
\bauthor{\bsnm{Cho}, \binits{M.}}:
\bctitle{Integralaction: Pose-driven feature integration for robust human
  action recognition in videos}.
In: \bbtitle{Proceedings of the IEEE/CVF Conference on Computer Vision and
  Pattern Recognition},
pp. \bfpage{3339}--\blpage{3348}
(\byear{2021})
\end{bchapter}
\endbibitem

%%% 74
\bibitem{zmigrod2013feature}
\begin{barticle}
\bauthor{\bsnm{Zmigrod}, \binits{S.}},
\bauthor{\bsnm{Hommel}, \binits{B.}}:
\batitle{Feature integration across multimodal perception and action: a
  review}.
\bjtitle{Multisensory research}
\bvolume{26}(\bissue{1-2}),
\bfpage{143}--\blpage{157}
(\byear{2013})
\end{barticle}
\endbibitem

%%% 75
\bibitem{levi1997feature}
\begin{barticle}
\bauthor{\bsnm{Levi}, \binits{D.M.}},
\bauthor{\bsnm{Sharma}, \binits{V.}},
\bauthor{\bsnm{Klein}, \binits{S.A.}}:
\batitle{Feature integration in pattern perception}.
\bjtitle{Proceedings of the National Academy of Sciences}
\bvolume{94}(\bissue{21}),
\bfpage{11742}--\blpage{11746}
(\byear{1997})
\end{barticle}
\endbibitem

%%% 76
\bibitem{peronamalikscale}
\begin{barticle}
\bauthor{\bsnm{Perona}, \binits{P.}},
\bauthor{\bsnm{Malik}, \binits{J.}}:
\batitle{Scale-space and edge detection using anisotropic diffusion}.
\bjtitle{IEEE Transactions on pattern analysis and machine intelligence}
\bvolume{12}(\bissue{7}),
\bfpage{629}--\blpage{639}
(\byear{1990})
\end{barticle}
\endbibitem

%%% 77
\bibitem{koenderink1984structure}
\begin{barticle}
\bauthor{\bsnm{Koenderink}, \binits{J.J.}}:
\batitle{The structure of images}.
\bjtitle{Biological cybernetics}
\bvolume{50}(\bissue{5}),
\bfpage{363}--\blpage{370}
(\byear{1984})
\end{barticle}
\endbibitem

%%% 78
\bibitem{koenderink2021structure}
\begin{barticle}
\bauthor{\bsnm{Koenderink}, \binits{J.}}:
\batitle{The structure of images: 1984--2021}.
\bjtitle{Biological cybernetics}
\bvolume{115}(\bissue{2}),
\bfpage{117}--\blpage{120}
(\byear{2021})
\end{barticle}
\endbibitem

%%% 79
\bibitem{noury2020deep}
\begin{botherref}
\oauthor{\bsnm{Noury}, \binits{Z.}},
\oauthor{\bsnm{Rezaei}, \binits{M.}}:
Deep-captcha: a deep learning based captcha solver for vulnerability
  assessment.
arXiv preprint arXiv:2006.08296
(2020)
\end{botherref}
\endbibitem

%%% 80
\bibitem{lin2018chinese}
\begin{barticle}
\bauthor{\bsnm{Lin}, \binits{D.}},
\bauthor{\bsnm{Lin}, \binits{F.}},
\bauthor{\bsnm{Lv}, \binits{Y.}},
\bauthor{\bsnm{Cai}, \binits{F.}},
\bauthor{\bsnm{Cao}, \binits{D.}}:
\batitle{Chinese character captcha recognition and performance estimation via
  deep neural network}.
\bjtitle{Neurocomputing}
\bvolume{288},
\bfpage{11}--\blpage{19}
(\byear{2018})
\end{barticle}
\endbibitem

%%% 81
\bibitem{liu2018towards}
\begin{bchapter}
\bauthor{\bsnm{Liu}, \binits{X.}},
\bauthor{\bsnm{Cheng}, \binits{M.}},
\bauthor{\bsnm{Zhang}, \binits{H.}},
\bauthor{\bsnm{Hsieh}, \binits{C.-J.}}:
\bctitle{Towards robust neural networks via random self-ensemble}.
In: \bbtitle{Proceedings of the European Conference on Computer Vision (ECCV)},
pp. \bfpage{369}--\blpage{385}
(\byear{2018})
\end{bchapter}
\endbibitem

%%% 82
\bibitem{frank2017validating}
\begin{barticle}
\bauthor{\bsnm{Frank}, \binits{M.R.}},
\bauthor{\bsnm{Cebrian}, \binits{M.}},
\bauthor{\bsnm{Pickard}, \binits{G.}},
\bauthor{\bsnm{Rahwan}, \binits{I.}}:
\batitle{Validating bayesian truth serum in large-scale online human
  experiments}.
\bjtitle{PloS one}
\bvolume{12}(\bissue{5}),
\bfpage{0177385}
(\byear{2017})
\end{barticle}
\endbibitem

%%% 83
\bibitem{medianabsolutedev}
\begin{barticle}
\bauthor{\bsnm{Rousseeuw}, \binits{P.J.}},
\bauthor{\bsnm{Croux}, \binits{C.}}:
\batitle{Alternatives to the median absolute deviation}.
\bjtitle{Journal of the American Statistical association}
\bvolume{88}(\bissue{424}),
\bfpage{1273}--\blpage{1283}
(\byear{1993})
\end{barticle}
\endbibitem

%%% 84
\bibitem{seabold2010statsmodels}
\begin{bchapter}
\bauthor{\bsnm{Seabold}, \binits{S.}},
\bauthor{\bsnm{Perktold}, \binits{J.}}:
\bctitle{Statsmodels: Econometric and statistical modeling with python}.
In: \bbtitle{9th Python in Science Conference}
(\byear{2010})
\end{bchapter}
\endbibitem

%%% 85
\bibitem{xie2020self}
\begin{bchapter}
\bauthor{\bsnm{Xie}, \binits{Q.}},
\bauthor{\bsnm{Luong}, \binits{M.-T.}},
\bauthor{\bsnm{Hovy}, \binits{E.}},
\bauthor{\bsnm{Le}, \binits{Q.V.}}:
\bctitle{Self-training with noisy student improves imagenet classification}.
In: \bbtitle{Proceedings of the IEEE/CVF Conference on Computer Vision and
  Pattern Recognition},
pp. \bfpage{10687}--\blpage{10698}
(\byear{2020})
\end{bchapter}
\endbibitem

%%% 86
\bibitem{liu2022towards}
\begin{bchapter}
\bauthor{\bsnm{Liu}, \binits{X.}},
\bauthor{\bsnm{Li}, \binits{W.}},
\bauthor{\bsnm{Yang}, \binits{Q.}},
\bauthor{\bsnm{Li}, \binits{B.}},
\bauthor{\bsnm{Yuan}, \binits{Y.}}:
\bctitle{Towards robust adaptive object detection under noisy annotations}.
In: \bbtitle{Proceedings of the IEEE/CVF Conference on Computer Vision and
  Pattern Recognition},
pp. \bfpage{14207}--\blpage{14216}
(\byear{2022})
\end{bchapter}
\endbibitem

%%% 87
\bibitem{chen2021robust}
\begin{bchapter}
\bauthor{\bsnm{Chen}, \binits{X.}},
\bauthor{\bsnm{Xie}, \binits{C.}},
\bauthor{\bsnm{Tan}, \binits{M.}},
\bauthor{\bsnm{Zhang}, \binits{L.}},
\bauthor{\bsnm{Hsieh}, \binits{C.-J.}},
\bauthor{\bsnm{Gong}, \binits{B.}}:
\bctitle{Robust and accurate object detection via adversarial learning}.
In: \bbtitle{Proceedings of the IEEE/CVF Conference on Computer Vision and
  Pattern Recognition},
pp. \bfpage{16622}--\blpage{16631}
(\byear{2021})
\end{bchapter}
\endbibitem

%%% 88
\bibitem{shen2020noise}
\begin{bchapter}
\bauthor{\bsnm{Shen}, \binits{Y.}},
\bauthor{\bsnm{Ji}, \binits{R.}},
\bauthor{\bsnm{Chen}, \binits{Z.}},
\bauthor{\bsnm{Hong}, \binits{X.}},
\bauthor{\bsnm{Zheng}, \binits{F.}},
\bauthor{\bsnm{Liu}, \binits{J.}},
\bauthor{\bsnm{Xu}, \binits{M.}},
\bauthor{\bsnm{Tian}, \binits{Q.}}:
\bctitle{Noise-aware fully webly supervised object detection}.
In: \bbtitle{Proceedings of the IEEE/CVF Conference on Computer Vision and
  Pattern Recognition},
pp. \bfpage{11326}--\blpage{11335}
(\byear{2020})
\end{bchapter}
\endbibitem

%%% 89
\bibitem{kaneko2020noise}
\begin{bchapter}
\bauthor{\bsnm{Kaneko}, \binits{T.}},
\bauthor{\bsnm{Harada}, \binits{T.}}:
\bctitle{Noise robust generative adversarial networks}.
In: \bbtitle{Proceedings of the IEEE/CVF Conference on Computer Vision and
  Pattern Recognition},
pp. \bfpage{8404}--\blpage{8414}
(\byear{2020})
\end{bchapter}
\endbibitem

%%% 90
\bibitem{munoz2010hymenoptera}
\begin{barticle}
\bauthor{\bsnm{Munoz-Torres}, \binits{M.C.}},
\bauthor{\bsnm{Reese}, \binits{J.T.}},
\bauthor{\bsnm{Childers}, \binits{C.P.}},
\bauthor{\bsnm{Bennett}, \binits{A.K.}},
\bauthor{\bsnm{Sundaram}, \binits{J.P.}},
\bauthor{\bsnm{Childs}, \binits{K.L.}},
\bauthor{\bsnm{Anzola}, \binits{J.M.}},
\bauthor{\bsnm{Milshina}, \binits{N.}},
\bauthor{\bsnm{Elsik}, \binits{C.G.}}:
\batitle{Hymenoptera genome database: integrated community resources for insect
  species of the order hymenoptera}.
\bjtitle{Nucleic acids research}
\bvolume{39}(\bissue{suppl\_1}),
\bfpage{658}--\blpage{662}
(\byear{2010})
\end{barticle}
\endbibitem

%%% 91
\bibitem{elsik2016hymenoptera}
\begin{barticle}
\bauthor{\bsnm{Elsik}, \binits{C.G.}},
\bauthor{\bsnm{Tayal}, \binits{A.}},
\bauthor{\bsnm{Diesh}, \binits{C.M.}},
\bauthor{\bsnm{Unni}, \binits{D.R.}},
\bauthor{\bsnm{Emery}, \binits{M.L.}},
\bauthor{\bsnm{Nguyen}, \binits{H.N.}},
\bauthor{\bsnm{Hagen}, \binits{D.E.}}:
\batitle{Hymenoptera genome database: integrating genome annotations in
  hymenopteramine}.
\bjtitle{Nucleic acids research}
\bvolume{44}(\bissue{D1}),
\bfpage{793}--\blpage{800}
(\byear{2016})
\end{barticle}
\endbibitem

\end{thebibliography}

%% if required, the content of .bbl file can be included here once bbl is generated
%%\input sn-article.bbl

%% Default %%
%%\input sn-sample-bib.tex%

\end{document}